\documentclass[10pt,twocolumn,letterpaper]{article}

\usepackage{cvpr}              

\usepackage{amsthm}
\newtheorem{proposition}{Proposition}
\usepackage{dsfont}
\usepackage[table]{xcolor}
\definecolor{plumhl}{RGB}{232, 222, 248}

\definecolor{cvprblue}{rgb}{0.21,0.49,0.74}
\usepackage[pagebackref,breaklinks,colorlinks,allcolors=cvprblue]{hyperref}

\def\confName{CVPR}
\def\confYear{2027}

\title{Box-conditional Hypothesis Testing for Out-of-Distribution Object Detection}

\author{Donghoon Lee, Jusung Park, Shinjin Kang\thanks{corresponding author}\\
School of Games, Hongik University
}

\begin{document}
\maketitle
\begin{abstract}
Detecting out-of-distribution (OOD) objects is essential for ensuring the reliability and safety of object detectors in real-world deployment. We treat the decision on a detected box as a hypothesis test with the null hypothesis that the box covers an object of an in-distribution (ID) class. The null distribution changes with the size of the box and its contact with the image border, and the detector observes both. The test can therefore rank a box only against ID boxes of the same size and border contact, which gives a distribution-free bound on the false-alarm rate for each of them and uncovers the separation that one null distribution hides. The power of the test also rests on the layer it reads. Layers built only for the ID classes have no reason to keep what sets an unknown object apart, so the test reads the features before them. We propose \textbf{PLUM}, a training-free score that pools the last backbone stage inside the box and ranks its distance to the ID classes among ID boxes of the same size and border contact. On the standard benchmark, PLUM reaches the highest AUROC among the compared methods without external models, with 91.5 and 94.5 on PASCAL-VOC and 95.1 and 97.5 on BDD100K.
Code: \url{https://github.com/DDOZMI/PLUM.git}
\end{abstract}

\section{Introduction}
\label{sec:intro}

Deep neural networks reach high accuracy when test-time data resemble the data they were trained on.
Deployed systems cannot count on this.
A network that receives an input unlike its training data still returns an answer, and it often returns this answer with high confidence~\cite{hendrycks2017msp}.
A system that relies on such answers therefore needs to know when an input lies outside what the network has learned.
Out-of-distribution (OOD) detection addresses this need.
An OOD detector gives each input a score that indicates how well the input fits the in-distribution (ID) data, and the system rejects inputs with low scores.
Most of this work studies image classification, where one decision covers the whole image.
Common scores use the confidence of the classifier~\cite{hendrycks2017msp,liang2018odin,liu2020energy} or the distance of its features to the ID data~\cite{lee2018mahalanobis,sun2022knn,wang2022vim}.

Object detection needs this decision for each object in an image.
A detector learns a closed set of classes~\cite{ren2015faster,zhu2021deformable}, yet the world contains many more classes of objects.
When a detector meets an object outside its classes, it still draws a box and assigns it an ID class, often with high confidence~\cite{du2022vos}.
OOD object detection therefore asks for every detected object whether it belongs to an ID class~\cite{du2022vos}.
Existing methods retrain the detector with synthetic outliers~\cite{du2022vos,du2022siren}, train a module on its features~\cite{wilson2023safe} or consult large vision-language models~\cite{zhang2025runa,li2026unioood}.

A detector judges an object through a box.
It places the box on a region of the image and pools features from inside the box.
The box therefore comes with more than the object.
It has a size and a position, and the image border may cut it.
It lies in an image with its own scene and photographic style.
It also rarely fits the object exactly, because the detector estimates its coordinates.
The decision concerns the object, yet the detector observes the box.
We treat the OOD decision on a detected box as a hypothesis test.
The null hypothesis states that the box covers an ID object, and the alternative states that it covers an object of an unknown class.
Each part of this statement places a demand on the test.

The first part is the box.
A car cut by the image border is still a car, and a giraffe in a small box is still a giraffe.
The scores of ID objects still change with the size of the box and with border contact, so the null distribution differs between boxes of different size and border contact.
We call the boxes of one size level and border contact a stratum.
The detector knows the stratum of each box at test time.
The test can therefore compare a box only with ID boxes of the same stratum.
A small box meets other small boxes, and a box at the image border meets other boxes at the border.
The second part is the object.
The test statistic must describe the object inside the box.
It should therefore read deep features pooled inside the box, because early layers describe local patterns and the whole image describes the scene.
The third part is the ID data.
An unknown object plays no role in the task the detector learns, and representations trained for a task are free to discard information that does not serve it~\cite{liu2020simple,vanamersfoort2020uncertainty,postels2020hidden}.
The feature pyramid and the box head exist only for this task, so the test should read the features before them.
These two demands meet at the last backbone stage, the deepest layer before the task-specific ones.

We propose \textbf{PLUM}, a \textbf{P}ooled, \textbf{L}eveled, \textbf{U}nit-norm Mahalanobis score.
PLUM pools the last backbone stage inside the box and measures the distance of this feature to ID statistics.
It then levels the distance by ranking each box only against ID boxes of the same stratum.
PLUM needs no training, no outlier data, no external model and no extra forward pass.
Ranking within strata and reading the last backbone stage each bring a large gain in separation.
False alarms stay near the chosen level in every stratum.
On PASCAL-VOC and BDD100K with Faster R-CNN, PLUM reaches the highest AUROC among the compared methods without external models, and the same recipe applies to Deformable DETR.

Our contributions are as follows.
\begin{itemize}
\item We cast the OOD decision on a detected box as a hypothesis test with the null hypothesis that the box covers an ID object. Conditioning this test on the observed stratum of the box gives a distribution-free bound on the false-alarm rate in every stratum. Its AUROC then depends on two choices, the null distribution and the representation that the score reads.
\item We define the task boundary of a detector, the point where the layers built only for the ID classes begin. Distances on the features at this boundary separate unknown objects best, and their power falls quickly past it in both Faster R-CNN and Deformable DETR.
\item We propose PLUM, a training-free OOD score for detected boxes. It reads the last backbone stage inside the box and ranks each box only against ID boxes of the same stratum.
\end{itemize}

\section{Related Work}
\label{sec:related}

\paragraph{OOD detection.}
An OOD score compares an input with the ID data that a model has learned.
Output-based scores read the confidence of the model, as in the maximum softmax probability~\cite{hendrycks2017msp}, ODIN~\cite{liang2018odin} and the energy score~\cite{liu2020energy}.
Feature-based scores measure how far a representation lies from the ID features.
They use the Mahalanobis distance~\cite{lee2018mahalanobis}, Gram matrices~\cite{sastry2020gram}, nearest neighbors~\cite{sun2022knn} or the residual outside a principal subspace~\cite{wang2022vim}.
Feature-based scores depend in turn on which features they read.
Normalizing features to unit length improves Mahalanobis-based scores~\cite{mueller2025mahapp}.
Classifiers trained without constraints can lose the information that sets OOD inputs apart, so one line of work constrains training with spectral normalization or a gradient penalty~\cite{liu2020simple,vanamersfoort2020uncertainty}.
Multilayer scores instead combine intermediate layers with the last layer after training~\cite{hossain2026mmpp}.
Work on neural collapse points the other way, because a linear probe trained on a layer detects OOD inputs better as that layer collapses more strongly~\cite{harun2025controlling}.
Any of these scores becomes a hypothesis test when ranked against held-out ID data, and the resulting conformal p-value bounds the false-alarm rate~\cite{haroush2022statistical,bates2023testing}.
These studies consider classifiers, where one feature describes the whole input and every input comes from a single ID distribution.

\paragraph{OOD object detection.}
OOD object detection carries such scores over to each detected object.
Early work showed that detectors assign ID classes to unknown objects~\cite{dhamija2020overlooked} and estimated their uncertainty with Monte-Carlo dropout~\cite{miller2018dropout}.
VOS~\cite{du2022vos} set the standard benchmark with PASCAL-VOC and BDD100K as ID data and MS-COCO and OpenImages as OOD data, and it retrains the detector with synthesized outliers.
Later methods change the synthesis or the representation~\cite{du2022siren,kumar2023ffs,wu2023srvae,wu2024tib,wu2023dfdd,wu2025wfs}.
SAFE~\cite{wilson2023safe} keeps the detector frozen and trains a small network on the backbone layers that are most sensitive to OOD inputs, and SyncOOD~\cite{liu2024syncood} trains on outliers from generative models.
Another line queries large pretrained models at test time~\cite{zhang2025runa,li2026unioood,peng2026unoadapter,nguyen2026ronin,gornhardt2026sam3o2d2}.
Scores that use only the frozen detector mostly serve as baselines on its classification branch~\cite{du2022vos}.
Most of these methods report AUROC and FPR95 pooled over all detections of the benchmark.

\paragraph{What a detection box carries.}
Several studies ask what such numbers reflect.
Images carry signatures of the dataset they come from~\cite{torralba2011unbiased,liu2025decade}, and recognition models draw on context and background~\cite{singh2020context,xiao2021noise}.
A model can then solve a benchmark through such shortcuts instead of the intended task~\cite{geirhos2020shortcut}.
Recent image-level OOD benchmarks therefore separate semantic novelty from covariate and dataset shift~\cite{yang2024imagenetood,bitterwolf2023ninco,zhang2023openood15}.
In the detection benchmark the ID and OOD images come from different datasets.
Analyses of this benchmark remove semantic overlap and add open-set metrics~\cite{montoya2025fmiyc}, count ID objects inside the OOD sets~\cite{he2025yolohallucination} and find that the advantage of VOS largely disappears when ID and OOD objects both come from MS-COCO~\cite{caldarella2025incremental}.
The box itself also shapes the output of a detector.
A fully connected box head is more sensitive to the spatial layout inside the box than a convolutional head, and it tells a complete object from a part of it~\cite{wu2020rethinking}.
The confidence of a detector depends on the size and the position of the box~\cite{kuppers2020multivariate,gilg2023boxsize}, and calibrating it within box sizes improves detection~\cite{gilg2023boxsize}.
The analyses revise the data and the metrics of the benchmark, and the calibration targets the probability that a detection is correct.

\section{OOD Decisions on Boxes}
\label{sec:formulation}

We treat the OOD decision on a detected box as a hypothesis test.
The null hypothesis states that the box covers an object of an ID class, and the alternative states that it covers an object of any other class.
In image classification one null distribution serves every input.
For detected boxes the null distribution moves with the box, and the box shows more than the object.
The null hypothesis names a box, an object and a set of ID classes, and each of them places a demand on the test.

\subsection{Setting}
\label{sec:setting}

Let $f$ be a trained object detector that we keep frozen, and let $\mathcal{C} = \{1, \dots, K\}$ be its classes.
For an image $x$, the detector returns detections, each with a box, a predicted class $c \in \mathcal{C}$ and a confidence $p$.
We write $b$ for the box from which the detector reads the object.
A detection is ID when this box covers an object of a class in $\mathcal{C}$ and OOD when it covers an object of any other class.
A score $S(x, b)$ is high for ID detections, and we write $S(b)$ when the image is clear from the context.

A box observation has four parts.
The object is the instance that the box covers, and its class decides the label.
The image is the whole image that contains the box, together with its scene and the dataset it comes from.
The geometry is the size of the box and its contact with the image border.
The placement is the offset between the box and the object.
Only the geometry follows from the box itself.
The source of the image and the placement remain unknown when the detector runs.

The test compares a score with a null distribution.
Let $F$ be the distribution function of the scores of ID boxes that serve as the reference.
The rank of a box is $u(b) = F(S(b))$, and the test flags the box as OOD when this rank is small,
\begin{equation}
  \text{OOD}(x, b) = \mathds{1}\left[\, u(b) \le \alpha \,\right].
  \label{eq:decision}
\end{equation}
When ID boxes have uniform ranks, the benchmark metrics describe this test directly,
\begin{equation}
  \begin{aligned}
    \mathrm{AUROC} &= 1 - \mathbb{E}_{\mathrm{OOD}}\left[ u(b) \right], \\
    \mathrm{FPR95} &= 1 - P_{\mathrm{OOD}}\left( u(b) \le 0.05 \right).
  \end{aligned}
  \label{eq:power}
\end{equation}
The AUROC is one minus the mean rank of OOD boxes, and the FPR95 is one minus the power of the test at level 0.05.
Both values rest on two choices, the null distribution $F$ and the representation that the score $S$ reads.

\subsection{The box}
\label{sec:box}

\begin{figure}[t]
  \centering
  \includegraphics[width=\linewidth]{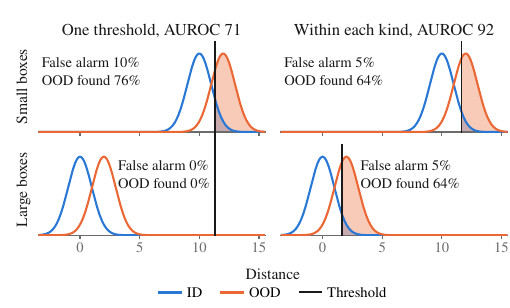}
  \caption{A simple example with equal numbers of small and large boxes. The distance follows a Gaussian with unit variance, with ID mean 10 on small boxes and 0 on large boxes, and OOD boxes lie 2 farther out. Each threshold keeps 95\% of the ID boxes, and shading marks the part of each distribution that the test flags as OOD. One threshold for all boxes reaches an AUROC of 71, and one threshold for each size reaches 92.}
  \label{fig:toy}
\end{figure}

The outputs of a detector depend on the size and the position of the box~\cite{kuppers2020multivariate,gilg2023boxsize,timans2024adaptive}.
One null distribution for all boxes then compares boxes of different sizes.
Figure~\ref{fig:toy} shows the effect in a simple example where OOD boxes lie equally far from ID boxes for small and large boxes alike, so the separation within each size is the same.
One threshold for all boxes rejects ID boxes unevenly and finds no OOD box among the large ones.
Comparing each box only with boxes of its own size evens out the false alarms and raises the AUROC from 71 to 92.
The uneven false alarms and the lost AUROC have one cause.
A conditional bias harms the ranking of detection confidences in a similar way~\cite{gilg2023boxsize}.

The size and border contact of a box are known at test time, so the test can condition on them.
We describe the geometry by the pair
\begin{equation}
  g(b) = \big(k(b), t(b)\big),
  \label{eq:geometry}
\end{equation}
where $k(b)$ is the feature-pyramid level of the box size~\cite{lin2017fpn} and $t(b)$ indicates whether the box touches the image border.
We call the set of boxes with one value of $g$ a stratum.
The test should keep the false-alarm rate in every stratum,
\begin{equation}
  P\big(\text{OOD}(x, b) = 1 \mid b \text{ is ID},\ g(b) = g\big) \le \alpha \quad \text{for every } g.
  \label{eq:req-geometry}
\end{equation}
We rank the score of a box only against the scores $r_1, \dots, r_n$ of held-out ID boxes from its own stratum,
\begin{equation}
  u(b) = \frac{1 + \left|\{\, i : r_i \le S(b) \,\}\right|}{n + 1}.
  \label{eq:rank}
\end{equation}

\begin{proposition}
\label{prop:validity}
Let $b$ be an ID box with $g(b) = g$, and let $r_1, \dots, r_n$ be the scores of held-out ID boxes from stratum $g$.
If the score of $b$ is exchangeable with $r_1, \dots, r_n$, then $P\big(u(b) \le \alpha \mid g(b) = g\big) \le \alpha$ for every $\alpha \in (0, 1)$.
\end{proposition}

The rank of $S(b)$ among the $n + 1$ exchangeable values is uniform, which gives the bound.
This construction is group-conditional conformal ranking~\cite{vovk2003mondrian,angelopoulos2024theoretical} with strata as the groups.
It meets Eq.~\eqref{eq:req-geometry} for any score and without any assumption on the distribution of the scores.
Detections in one image depend on each other, and reference boxes from the images on which the detector was trained need not be exchangeable with test boxes.

\begin{proposition}
\label{prop:shift}
Let $r_1, \dots, r_n$ be any reference scores of stratum $g$, and let $\hat F^-(t)$ and $G^-(t)$ be the share of these scores and the probability that the score of an ID box of stratum $g$ lies below $t$.
With $l_{\max} = \lfloor (n + 1)\alpha_{\max} \rfloor$, every $\alpha \le \alpha_{\max}$ satisfies $P\big(u(b) \le \alpha \mid g(b) = g\big) \le \alpha + \sup_{t \le r_{(l_{\max})}} \big(G^-(t) - \hat F^-(t)\big)$.
\end{proposition}

The bound needs no assumption on the references, and it also covers one null distribution, whose $\hat F^-$ holds references of every stratum.
We write $\delta$ for its second term with $\alpha_{\max}$ of 10\% and check the requirement on ID data only, by the false-alarm rate in each stratum and by $\delta$.
Ranks that are uniform within strata are also uniform overall, so Eq.~\eqref{eq:power} now counts each box against the null distribution of its own stratum.
Both bounds hold for any score, so the score may first remove what the geometry still explains within a stratum.

\subsection{The object}
\label{sec:object}

The test statistic must describe the object.
Recognition models draw on context and background~\cite{singh2020context,xiao2021noise}, and the ID and OOD images of the standard benchmark come from different datasets~\cite{du2022vos}.
A score that reads the image can therefore separate ID from OOD images without looking at the object.
On a box that contains no object, the score should not separate them.
For such a box $\bar{b}$, we require
\begin{equation}
  S(x, \bar{b}) \mid x \text{ is ID} \;\overset{d}{=}\; S(x, \bar{b}) \mid x \text{ is OOD},
  \label{eq:req-image}
\end{equation}
where $\overset{d}{=}$ denotes equality in distribution.
We check this requirement on background boxes, which are boxes that contain no object.
The background AUROC $A_{\mathrm{bg}}$ compares background boxes of ID images with those of OOD images.
Under Eq.~\eqref{eq:req-image}, it equals 50, because a background box from an ID image then scores higher with probability one half.
A departure in either direction shows that the score separates ID from OOD images without the object.

The detector estimates the coordinates of a box, so the placement is rarely zero.
Fully connected box heads are sensitive to the spatial layout inside the box~\cite{wu2020rethinking}, so their features may change with the placement while the object stays the same.
A small change of placement should change the score little.
For a box $b'$ that covers the same object as $b$ with a slightly different placement, we require $S(x, b') \approx S(x, b)$.
We check this requirement by perturbing the boxes slightly and measuring how much the AUROC changes.

Neither the source of the image nor the placement is known when the detector runs.
The test therefore cannot condition on them, and only the score can ignore them.
The score must also read a layer that describes objects.
Early layers describe local patterns rather than objects, so the statistic should read a deep layer.

\subsection{The ID classes}
\label{sec:idclasses}

The alternative hypothesis concerns objects outside the ID classes, and the detector never learns to tell these objects apart.
Representations trained for a task are free to discard information that does not serve it~\cite{liu2020simple,vanamersfoort2020uncertainty,postels2020hidden}.
We define the task boundary of a detector as the point where the layers built only for the ID classes begin.
In Faster R-CNN the boundary lies before the feature pyramid, and in Deformable DETR it lies before the transformer.
Layers past the boundary have no reason to preserve what sets an unknown object apart, so the test should read the features before them.
A linear probe trained on a later layer can still read what remains there~\cite{harun2025controlling}.
A distance to ID statistics needs no such training, and it dilutes evidence that a few directions hold.

The demand for depth and the task boundary meet at the last backbone stage, the deepest layer before the boundary.
The box thus fixes the null distribution $F$, and the object and the ID classes fix the layer that the score reads.
What remains is the distance on this layer.

\section{Method}
\label{sec:method}

\begin{figure*}[t]
  \centering
  \includegraphics[width=\textwidth]{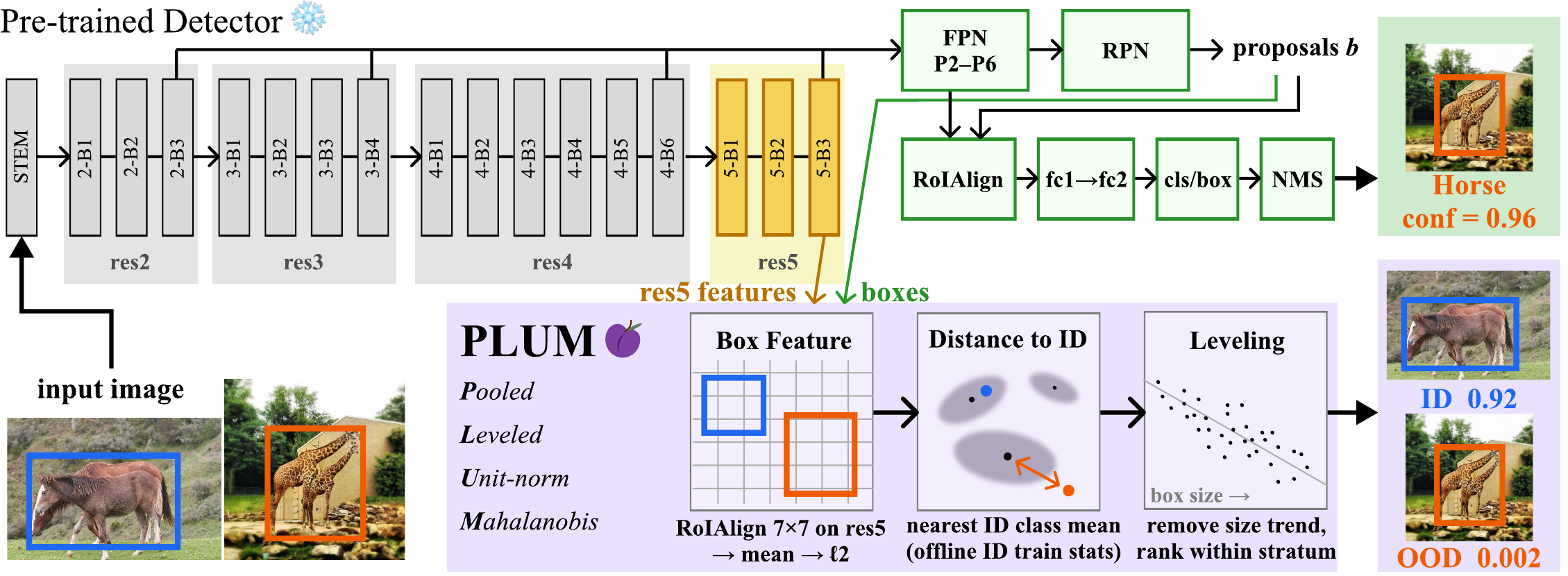}
  \caption{Overview of PLUM.
    The detector stays frozen.
    PLUM pools the res5 feature inside the box from which each detection is read and measures its Mahalanobis distance to the ID class means.
    It then ranks this distance among ID reference detections of the same stratum.
    A giraffe detected as a horse with confidence 0.96 receives a PLUM value of 0.002, whereas an ID horse receives 0.92.}
  \label{fig:overview}
\end{figure*}

PLUM consists of a score and a decision.
The score pools backbone features inside the box.
The decision ranks this score only against ID boxes of the same stratum.
We call the result PLUM, a Pooled, Leveled, Unit-norm Mahalanobis score.
Figure~\ref{fig:overview} gives an overview.

\subsection{Box feature}
\label{sec:method_feat}

PLUM reads each detection through the box $b$ from which the detector reads the object.
In a two-stage detector~\cite{ren2015faster}, this is the proposal box that the box head pools.
In a detector without proposals such as Deformable DETR~\cite{zhu2021deformable}, it is the output box.
Let $A(x) \in \mathbb{R}^{D \times H \times W}$ be the output of the last backbone stage, the deepest layer before the task boundary of Sec.~\ref{sec:idclasses}.
For a ResNet-50 backbone it is the res5 map with $D = 2048$ channels.
PLUM pools $A(x)$ over the box with RoIAlign~\cite{he2017maskrcnn} on a $7 \times 7$ grid and averages the grid cells,
\begin{equation}
  \phi(x, b) = \frac{1}{49} \sum_{i,j=1}^{7} \mathrm{RoIAlign}\big(A(x), b\big)_{ij}.
  \label{eq:pool}
\end{equation}
The grid matches the box head of the feature pyramid network~\cite{lin2017fpn}.
PLUM then normalizes the pooled feature to unit length~\cite{mueller2025mahapp},
\begin{equation}
  z(x, b) = \frac{\phi(x, b)}{\lVert \phi(x, b) \rVert_2}.
  \label{eq:feature}
\end{equation}
These steps serve the demands of Sec.~\ref{sec:object}.
Pooling inside the box keeps the feature on the object, and the average over the grid discards where the object sits inside the box.
The receptive field of the feature still reaches beyond the box, so pooling limits what the image adds without removing it.

\subsection{Distance to ID statistics}
\label{sec:method_dist}

PLUM takes its ID statistics from the detector itself.
It runs the detector on the ID training images with the same detection rule as at test time and keeps every detection with its box and its predicted class.
No ground-truth label enters the statistics.
Images with an even index give the class statistics, and we write $\mathcal{D}_{\mathrm{S}}$ for their detections.
For a detection with box $b$, we write $z_b$ for its feature from Eq.~\eqref{eq:feature} and $c_b$ for its predicted class.
Let $\mathcal{D}_{\mathrm{S}}^{c}$ be the detections in $\mathcal{D}_{\mathrm{S}}$ with $c_b = c$.
The class means are
\begin{equation}
  \mu_c = \frac{1}{|\mathcal{D}_{\mathrm{S}}^{c}|} \sum_{b \in \mathcal{D}_{\mathrm{S}}^{c}} z_b.
  \label{eq:mean}
\end{equation}
All classes share one covariance~\cite{lee2018mahalanobis}, which PLUM estimates from the scatter around the class means,
\begin{equation}
  \Sigma_0 = \frac{1}{|\mathcal{D}_{\mathrm{S}}|} \sum_{b \in \mathcal{D}_{\mathrm{S}}} (z_b - \mu_{c_b})(z_b - \mu_{c_b})^{\top}.
  \label{eq:scatter}
\end{equation}
The scatter of a high-dimensional feature is poorly conditioned.
PLUM shrinks it towards a scaled identity with the closed-form estimator of Ledoit and Wolf~\cite{ledoit2004wellconditioned},
\begin{equation}
  \hat{\Sigma} = (1 - \rho)\,\Sigma_0 + \rho\, m\, \mathbf{I}, \qquad m = \frac{\operatorname{tr} \Sigma_0}{D},
  \label{eq:lw}
\end{equation}
where the shrinkage weight follows from the same detections,
\begin{equation}
  \begin{split}
    \rho &= \frac{\min(\bar{\beta}^2, \delta^2)}{\delta^2}, \qquad \delta^2 = \lVert \Sigma_0 - m\,\mathbf{I} \rVert_F^2, \\
    \bar{\beta}^2 &= \frac{1}{|\mathcal{D}_{\mathrm{S}}|^2} \sum_{b \in \mathcal{D}_{\mathrm{S}}} \big\lVert (z_b - \mu_{c_b})(z_b - \mu_{c_b})^{\top} - \Sigma_0 \big\rVert_F^2.
  \end{split}
  \label{eq:lw-weight}
\end{equation}
The raw score of a box is the negative Mahalanobis distance to the nearest class mean,
\begin{equation}
  s(x, b) = -\min_{c \in \mathcal{C}}\, \big(z(x, b) - \mu_c\big)^{\top} \hat{\Sigma}^{-1} \big(z(x, b) - \mu_c\big).
  \label{eq:dist}
\end{equation}
The minimum runs over all classes, so the predicted class of a test detection does not enter the score.

\subsection{Leveling}
\label{sec:method_level}

PLUM levels the raw score within the strata of Sec.~\ref{sec:box}.
It computes the size level of a box with the level rule of the feature pyramid network~\cite{lin2017fpn},
\begin{equation}
  k(b) = \min\Big(5, \max\Big(2, \Big\lfloor 4 + \log_2 \tfrac{\sqrt{w_b h_b}}{224} \Big\rfloor\Big)\Big),
  \label{eq:level}
\end{equation}
where $w_b$ and $h_b$ are the width and height of the box in the network input.
The same rule applies to detectors without a feature pyramid.
We set $t(b) = 1$ when one side of the box lies on a side of the image.

\begin{table*}[t]
  \centering
  \caption{OOD object detection with a ResNet-50 Faster R-CNN. Bold marks the best value in each column. $^\dagger$Published MLP. $^\ddagger$Reported in~\cite{du2022vos}. SIREN reports Faster R-CNN results only on PASCAL-VOC.}
  \label{tab:main}
  \footnotesize
  \begin{tabular}{l c cccc cccc}
    \toprule
     & & \multicolumn{4}{c}{\textbf{ID: PASCAL-VOC}} & \multicolumn{4}{c}{\textbf{ID: BDD100K}} \\
    \cmidrule(lr){3-6} \cmidrule(lr){7-10}
     & & \multicolumn{2}{c}{MS-COCO} & \multicolumn{2}{c}{OpenImages} & \multicolumn{2}{c}{MS-COCO} & \multicolumn{2}{c}{OpenImages} \\
    \cmidrule(lr){3-4} \cmidrule(lr){5-6} \cmidrule(lr){7-8} \cmidrule(lr){9-10}
    Method & Retrain? & FPR95$\downarrow$ & AUROC$\uparrow$ & FPR95$\downarrow$ & AUROC$\uparrow$ & FPR95$\downarrow$ & AUROC$\uparrow$ & FPR95$\downarrow$ & AUROC$\uparrow$ \\
    \midrule
    MSP~\cite{hendrycks2017msp} &  & 72.04 & 83.20 & 74.57 & 81.44 & 80.77 & 75.85 & 79.04 & 77.10 \\
    ODIN~\cite{liang2018odin} &  & 57.73 & 83.25 & 59.47 & 82.64 & 62.80 & 74.43 & 59.45 & 76.20 \\
    Energy~\cite{liu2020energy} &  & 57.40 & 83.55 & 59.70 & 82.64 & 60.03 & 77.47 & 55.95 & 79.23 \\
    Mahalanobis~\cite{lee2018mahalanobis} &  & 96.88 & 57.98 & 97.56 & 55.40 & 56.97 & 88.09 & 59.45 & 87.06 \\
    KNN~\cite{sun2022knn} &  & 60.28 & 84.74 & 60.76 & 81.98 & 42.53 & 90.11 & 39.41 & 90.83 \\
    ViM~\cite{wang2022vim} &  & 92.02 & 74.61 & 92.54 & 72.24 & 45.80 & 90.44 & 46.95 & 90.66 \\
    Gram~\cite{sastry2020gram}$^\ddagger$ &  & 62.75 & 79.88 & 67.42 & 77.62 & 60.93 & 74.93 & 77.55 & 59.38 \\
    SAFE~\cite{wilson2023safe}$^\dagger$ &  & 51.07 & 78.61 & \textbf{24.04} & 90.71 & \textbf{23.85} & 92.02 & 13.95 & 95.28 \\
    \midrule
    G-ODIN~\cite{hsu2020godin}$^\ddagger$ & \checkmark & 59.57 & 83.12 & 70.28 & 79.23 & 57.27 & 85.22 & 50.17 & 87.18 \\
    CSI~\cite{tack2020csi}$^\ddagger$ & \checkmark & 59.91 & 81.83 & 57.41 & 82.95 & 47.10 & 84.09 & 37.06 & 87.99 \\
    GAN-synthesis~\cite{lee2018training}$^\ddagger$ & \checkmark & 60.93 & 83.67 & 59.97 & 82.67 & 57.03 & 78.82 & 50.61 & 81.25 \\
    VOS~\cite{du2022vos} & \checkmark & 46.35 & 89.61 & 54.44 & 85.70 & 45.88 & 86.50 & 37.00 & 88.73 \\
    SIREN~\cite{du2022siren} & \checkmark & 51.69 & 88.41 & 53.13 & 87.15 & -- & -- & -- & -- \\
    \midrule
    \textbf{PLUM (Ours)} &  & \textbf{44.16} & \textbf{91.46} & 29.25 & \textbf{94.50} & 24.35 & \textbf{95.07} & \textbf{13.87} & \textbf{97.51} \\
    \bottomrule
  \end{tabular}
\end{table*}

The images with an odd index take turns.
One half gives the detections $\mathcal{D}_{\mathrm{L}}$ that fit the lines below, and the other half gives the reference detections $\mathcal{D}_{\mathrm{R}}$.
Within a stratum the raw score still changes with the size of the box, so PLUM first removes this trend.
For each stratum $g$, it fits a line in log area to the raw scores of $\mathcal{D}_{\mathrm{L}}$,
\begin{equation}
  (\theta_g, \eta_g) = \operatorname*{arg\,min}_{\theta, \eta} \sum_{\substack{b \in \mathcal{D}_{\mathrm{L}} \\ g(b) = g}} \big(s_b - \theta - \eta \log a(b)\big)^2,
  \label{eq:ols}
\end{equation}
where $s_b$ is the raw score of the detection and $a(b) = w_b h_b$ is the area of its box.
The score of a box is its residual with respect to the line of its own stratum $g = g(b)$,
\begin{equation}
  S(x, b) = s(x, b) - \theta_g - \eta_g \log a(b).
  \label{eq:resid}
\end{equation}
Let $\mathcal{D}_{\mathrm{R}}^{g}$ be the detections in $\mathcal{D}_{\mathrm{R}}$ with $g(b) = g$, and let $S_{b'}$ be the score of a reference detection $b'$.
The output of PLUM ranks this score among the reference scores of the same stratum,
\begin{equation}
  u(x, b) = \frac{1 + \big|\{\, b' \in \mathcal{D}_{\mathrm{R}}^{g} : S_{b'} \le S(x, b) \,\}\big|}{|\mathcal{D}_{\mathrm{R}}^{g}| + 1}.
  \label{eq:plum}
\end{equation}
This is the rank of Eq.~\eqref{eq:rank}.
The reference detections come from images that neither the class statistics nor the lines use, so Proposition~\ref{prop:validity} covers PLUM in every stratum as far as these training detections are exchangeable with test boxes.
A low value means that few ID boxes of the same stratum score lower.

PLUM leaves the detector and its detections unchanged, so the ID detection accuracy stays that of the detector.
It trains no module and uses neither outlier data nor any model other than the detector.
A single forward pass suffices because the detector computes $A(x)$ anyway.
All statistics have closed forms, namely the class means, one covariance, two line coefficients per stratum and the sorted reference residuals.
The task boundary fixes the layer, the grid and the size levels follow the design of the feature pyramid network, and no part of PLUM is set on OOD data.

\section{Experiments}
\label{sec:experiments}

\subsection{Experimental setup}
\label{sec:setup}

\paragraph{Benchmark.}
We follow the standard benchmark~\cite{du2022vos}.
The ID datasets are PASCAL-VOC and BDD100K, and the OOD datasets are images from MS-COCO and OpenImages that contain no object of an ID class.
We evaluate a detection when its maximum foreground probability exceeds a threshold chosen on the ID validation set, which is 0.5728 for PASCAL-VOC and 0.4006 for BDD100K.
Every evaluated detection on an ID image counts as ID and every evaluated detection on an OOD image counts as OOD.
This gives 17{,}155 ID detections against 1{,}216 and 2{,}171 OOD detections on PASCAL-VOC, and 206{,}339 ID detections against 2{,}382 and 1{,}312 on BDD100K.

\paragraph{Metrics.}
We report FPR95 and AUROC with ID as the positive class.
By Eq.~\eqref{eq:power}, both describe the power of the test, so we also report the false-alarm rate that the test holds at a given level.
Together these are the size and the power of the test.

\paragraph{Detectors.}
The main detector is the public Faster R-CNN~\cite{ren2015faster} with a ResNet-50 feature pyramid released with the benchmark, trained on the ID dataset without any OOD-specific objective.
Sec.~\ref{sec:exp_power} also reads the layers of Deformable DETR~\cite{zhu2021deformable}.
The supplementary material reports the full results with Deformable DETR.

\paragraph{Compared methods.}
Scores that leave the detector unchanged run on the same detector and the same detections as PLUM.
Methods that retrain the detector run from their public checkpoints with their own detections and thresholds.
For methods without a public detection implementation, we copy the values reported with the benchmark.

\paragraph{Implementation.}
The ID statistics of PLUM come from the detections on all PASCAL-VOC training images and on 10{,}000 randomly drawn BDD100K training images.
We compute distances in double precision.

\begin{table}[t]
  \centering
  \caption{Size and power of the res5 distance under different null distributions, averaged over MS-COCO and OpenImages as OOD sets. False alarms give the range over strata of the share of ID boxes with rank at most 5\%.}
  \label{tab:null}
  \footnotesize
  \setlength{\tabcolsep}{2.5pt}
  \resizebox{\linewidth}{!}{%
  \begin{tabular}{l ccc ccc}
    \toprule
     & \multicolumn{3}{c}{PASCAL-VOC} & \multicolumn{3}{c}{BDD100K} \\
    \cmidrule(lr){2-4} \cmidrule(lr){5-7}
    Null distribution & AUROC$\uparrow$ & FPR95$\downarrow$ & False alarms & AUROC$\uparrow$ & FPR95$\downarrow$ & False alarms \\
    \midrule
    One for all boxes & 74.2 & 80.2 & 0.05--20.1 & 94.4 & 38.8 & 0.0--36.6 \\
    Size level & 91.8 & 40.7 & 2.5--6.3 & 96.2 & 19.3 & 3.4--36.2 \\
    Size level $\times$ border & 91.8 & 41.1 & 0.8--7.4 & 95.9 & 22.0 & 2.6--5.3 \\
    PLUM, with line & \textbf{93.0} & \textbf{36.7} & 0.8--7.5 & \textbf{96.3} & \textbf{19.1} & 2.6--5.8 \\
    \bottomrule
  \end{tabular}%
  }
\end{table}

\subsection{Power of the test}
\label{sec:exp_power}

\begin{table}[t]
  \centering
  \caption{The PLUM distance read at different layers, averaged over MS-COCO and OpenImages as OOD sets. Raw uses one null distribution and Leveled applies Sec.~\ref{sec:method_level}. Rules mark the task boundary, and bold marks the best Leveled value of each detector. BDD100K has no feature-pyramid dump.}
  \label{tab:layers}
  \footnotesize
  \setlength{\tabcolsep}{2.5pt}
  \resizebox{\linewidth}{!}{%
  \begin{tabular}{l cccc cccc}
    \toprule
     & \multicolumn{4}{c}{PASCAL-VOC} & \multicolumn{4}{c}{BDD100K} \\
    \cmidrule(lr){2-5} \cmidrule(lr){6-9}
     & \multicolumn{2}{c}{AUROC$\uparrow$} & \multicolumn{2}{c}{FPR95$\downarrow$} & \multicolumn{2}{c}{AUROC$\uparrow$} & \multicolumn{2}{c}{FPR95$\downarrow$} \\
    \cmidrule(lr){2-3} \cmidrule(lr){4-5} \cmidrule(lr){6-7} \cmidrule(lr){8-9}
    Layer & Raw & Leveled & Raw & Leveled & Raw & Leveled & Raw & Leveled \\
    \midrule
    \multicolumn{9}{l}{\textit{Faster R-CNN}} \\
    res2 & 72.5 & 86.8 & 82.6 & 49.3 & 53.5 & 89.6 & 92.5 & 51.6 \\
    res3 & 66.9 & 88.4 & 93.6 & 47.3 & 51.8 & 92.3 & 93.7 & 38.9 \\
    res4 & 63.5 & 89.7 & 91.8 & 45.7 & 78.2 & 94.9 & 74.9 & 30.0 \\
    res5 & 74.2 & \textbf{93.0} & 80.2 & \textbf{36.7} & 94.4 & \textbf{96.3} & 38.8 & \textbf{19.1} \\
    \midrule
    FPN & 53.9 & 82.3 & 92.3 & 65.0 & -- & -- & -- & -- \\
    fc1 & 54.7 & 60.2 & 98.0 & 96.7 & 84.4 & 88.2 & 78.0 & 62.0 \\
    fc2 & 58.5 & 65.0 & 96.5 & 94.7 & 87.8 & 86.6 & 60.2 & 70.2 \\
    logit & 78.0 & 76.9 & 68.5 & 73.4 & 86.8 & 84.1 & 49.9 & 55.2 \\
    \midrule
    \multicolumn{9}{l}{\textit{Deformable DETR}} \\
    backbone & 80.4 & \textbf{92.9} & 75.4 & \textbf{34.4} & 95.8 & \textbf{94.4} & 24.6 & \textbf{28.3} \\
    \midrule
    decoder & 57.6 & 48.8 & 94.1 & 97.3 & 94.3 & 87.6 & 27.8 & 60.9 \\
    \bottomrule
  \end{tabular}%
  }
\end{table}

\subsection{Benchmark results}
\label{sec:benchmark}

\begin{figure*}[t]
  \centering
  \includegraphics[width=\textwidth]{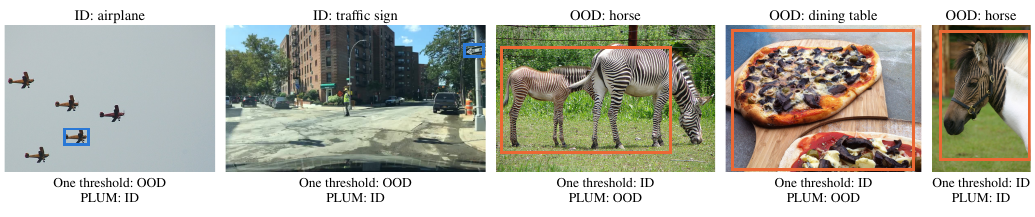}
  \caption{Detections on which the two tests differ, and one failure. Each panel gives the true status of the box and the class that Faster R-CNN predicts, and the test runs at $\alpha = 5\%$. These detections illustrate each case and are not a sample.}
  \label{fig:examples}
\end{figure*}

\paragraph{Comparison.}
Table~\ref{tab:main} reports the benchmark.
PLUM has the highest AUROC in all four pairings of ID and OOD datasets and leads the next method by 1.9 to 3.8 points.
Its FPR95 is the lowest in two pairings and stays within 0.5 points of the lowest with MS-COCO as OOD on BDD100K.
With OpenImages as OOD on PASCAL-VOC, its FPR95 is 29.25 against the lowest 24.04.
PLUM keeps the detections unchanged, so its ID detection accuracy is that of the detector with 48.55 and 31.24 COCO-style AP on PASCAL-VOC and BDD100K.

\paragraph{Qualitative results.}
Figure~\ref{fig:examples} shows the effect on single detections.
One threshold for all boxes rejects a small airplane and a traffic sign at the image border, and PLUM keeps both.
It also accepts the large boxes on zebras and on a pizza, and PLUM flags them.
The last panel shows a failure, where a zebra that resembles a horse passes both tests.

\subsection{Size of the test}
\label{sec:exp_size}

\begin{figure}[t]
  \centering
  \includegraphics[width=\linewidth]{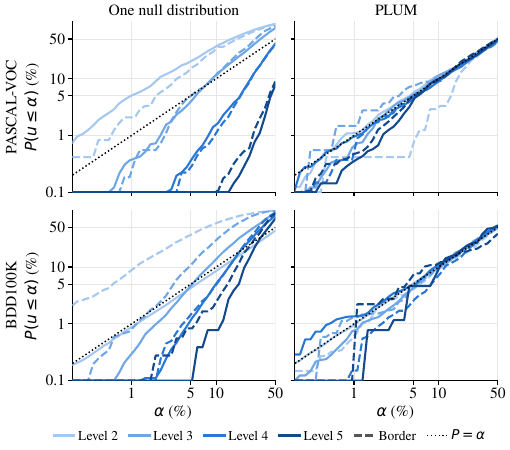}
  \caption{Share of ID validation boxes with rank at most $\alpha$ in each stratum, with a ResNet-50 Faster R-CNN. A valid test follows the dotted diagonal. Line color gives the feature-pyramid level of the stratum, and dashed lines mark boxes at the image border. Shares below 0.1\% are drawn at 0.1\%.}
  \label{fig:calib}
\end{figure}

\paragraph{False alarms at every level.}
A test at level $\alpha$ should flag at most a share $\alpha$ of the ID boxes of every stratum.
Figure~\ref{fig:calib} measures this share on the ID validation boxes.
One null distribution spreads the strata widely around the diagonal, just as in the example of Fig.~\ref{fig:toy}, and PLUM brings them onto it.
Its reference boxes come from the training images, and Table~\ref{tab:exch} measures $\delta$ of Proposition~\ref{prop:shift} on the ID validation boxes.
With one null distribution, $\delta$ reaches 21 to 49 points for small boxes and boxes at the border, and it stays at most 5.2 points with PLUM.
Against exchangeable references, $\delta$ of PLUM is significant at the 5\% level in two of 31 strata.

\paragraph{What the test conditions on.}
Size alone does not define the strata.
In Table~\ref{tab:null}, conditioning only on the size level still lets the false alarms of a stratum reach 36.2\% on BDD100K.
Adding border contact limits them to 5.3\% at little cost in AUROC.

\begin{table}[t]
  \centering
  \caption{Largest $\delta$ over strata in points, and strata where $\delta$ is significant at the 5\% level against exchangeable references.}
  \label{tab:exch}
  \footnotesize
  \setlength{\tabcolsep}{3pt}
  \resizebox{\linewidth}{!}{%
  \begin{tabular}{l l c cc cc}
    \toprule
     & & & \multicolumn{2}{c}{$\delta$} & \multicolumn{2}{c}{Significant} \\
    \cmidrule(lr){4-5} \cmidrule(lr){6-7}
    Detector & ID & Strata & One null & PLUM & One null & PLUM \\
    \midrule
    Faster R-CNN & PASCAL-VOC & 8 & 27.0 & 3.0 & 3 & 1 \\
     & BDD100K & 8 & 49.1 & 5.2 & 3 & 0 \\
    \midrule
    Deformable DETR & PASCAL-VOC & 8 & 21.0 & 0.1 & 2 & 0 \\
     & BDD100K & 7 & 22.7 & 2.1 & 2 & 1 \\
    \bottomrule
  \end{tabular}%
  }
\end{table}

\paragraph{Null distribution.}
We first keep the last backbone stage and change only the null distribution, as in Table~\ref{tab:null}.
Ranks keep the order of the boxes inside a stratum, so the gain of the first step comes only from comparisons between strata.
These comparisons make up 83.4\% and 83.7\% of the ID and OOD pairs on PASCAL-VOC with MS-COCO and OpenImages.

\paragraph{Layer.}
We then keep the distance and the ranking and change only the layer.
Table~\ref{tab:layers} shows the Leveled distance rising along the backbone and falling past the task boundary, in Faster R-CNN and in Deformable DETR.
Among the layers of Faster R-CNN, res5 has the best AUROC and FPR95 in all four pairings.
Ranking does not bring the evidence back past the boundary.

\paragraph{Both choices together.}
The two choices depend on each other.
On PASCAL-VOC, moving the raw distance from fc2 to res5 adds 15.7 points of AUROC and leveling fc2 adds 6.5, whereas both together add 34.5.
One null distribution also misjudges the layers, because it places the logits above res5 on PASCAL-VOC and ranking within strata reverses this order.

\paragraph{Power within strata.}
Equation~\eqref{eq:power} also holds within a stratum, so Table~\ref{tab:kinds} reads the power of PLUM in each stratum.
The smallest boxes on PASCAL-VOC have the lowest power, and boxes at the image border generally have less power than boxes of the same size away from it.
A box cut by the border shows only part of its object, which may leave less evidence for the test.

\subsection{What the test reads}
\label{sec:exp_object}

\paragraph{Image.}
For each detection we place a background box of the same size at a random position in the same image where it overlaps no detection, and we leave out detections without room for such a box.
On BDD100K and with MS-COCO as OOD on PASCAL-VOC, the background AUROC of PLUM stays within 6 points of 50.
With OpenImages as OOD on PASCAL-VOC, it reaches 73.5, so PLUM separates part of these images without an object.

\paragraph{Placement.}
We perturb every box with random offsets~\cite{gornhardt2026sam3o2d2}.
Each coordinate moves by an offset drawn uniformly within $\varepsilon / 2$ times the width or height of the box, and we clip the box to the image.
At $\varepsilon = 0.5$, the perturbed box overlaps the original one with a mean IoU of 0.64, and PLUM loses at most 0.6 points of AUROC.

\begin{table}[t]
  \centering
  \caption{AUROC of PLUM within each stratum against the ID boxes of the same stratum. The level gives the size of the box, and Border marks boxes that touch the image border. A dash marks strata with fewer than 30 OOD detections.}
  \label{tab:kinds}
  \footnotesize
  \setlength{\tabcolsep}{2.5pt}
  \resizebox{\linewidth}{!}{%
  \begin{tabular}{l cc cc cc cc}
    \toprule
     & \multicolumn{4}{c}{PASCAL-VOC} & \multicolumn{4}{c}{BDD100K} \\
    \cmidrule(lr){2-5} \cmidrule(lr){6-9}
     & \multicolumn{2}{c}{MS-COCO} & \multicolumn{2}{c}{OpenImages} & \multicolumn{2}{c}{MS-COCO} & \multicolumn{2}{c}{OpenImages} \\
    \cmidrule(lr){2-3} \cmidrule(lr){4-5} \cmidrule(lr){6-7} \cmidrule(lr){8-9}
    Level & Inner & Border & Inner & Border & Inner & Border & Inner & Border \\
    \midrule
    2 & 84.8 & -- & 92.8 & -- & 95.9 & 90.5 & 97.9 & 96.1 \\
    3 & 90.8 & -- & 94.5 & -- & 96.8 & 94.8 & 97.4 & -- \\
    4 & 94.2 & 88.6 & 95.4 & 90.6 & 96.8 & 96.8 & 98.0 & 98.0 \\
    5 & 92.9 & 92.7 & 95.1 & 94.6 & 93.8 & 91.0 & 97.9 & 96.9 \\
    \bottomrule
  \end{tabular}%
  }
\end{table}
\section{Conclusion}
\label{sec:conclusion}

We wrote the OOD decision on a detected box as a hypothesis test with the null hypothesis that the box covers an object of an ID class.
This view assigns the behavior of a score to two choices.
Ranking within strata keeps the false-alarm rate in every stratum and uncovers the separation that one null distribution hides.
Reading the features before the task boundary keeps this separation, and the power of the test falls past the boundary in both Faster R-CNN and Deformable DETR.
From these choices we derived PLUM, a score that needs no training and runs on a frozen detector.

Its bound needs reference boxes that are exchangeable with test boxes. PLUM takes them from the training images and treats the boxes of one image as independent.

Treating each box as a test opens object detection to the tools of statistical inference, starting with the boxes of one image.
Multiple testing can control the errors over all detections in an image~\cite{bates2023testing,ma2025online}, and evidence can accumulate over the frames of a video.
The same test applies to other recognition tasks that observe an object through a unit such as a mask or a 3D box.

{
    \small
    \bibliographystyle{ieeenat_fullname}
    \bibliography{main}

@String(ICLR = {Int. Conf. Learn. Represent.})

@String(AAAI = {AAAI})

@String(ICLR  = {ICLR})

@inproceedings{hendrycks2017msp,
  author    = {Dan Hendrycks and Kevin Gimpel},
  title     = {A Baseline for Detecting Misclassified and Out-of-Distribution Examples in Neural Networks},
  booktitle = {International Conference on Learning Representations},
  year      = {2017},
  eprint    = {1610.02136},
  archivePrefix = {arXiv}
}

@inproceedings{liang2018odin,
  author    = {Shiyu Liang and Yixuan Li and R. Srikant},
  title     = {Enhancing The Reliability of Out-of-distribution Image Detection in Neural Networks},
  booktitle = {International Conference on Learning Representations},
  year      = {2018},
  url       = {https://openreview.net/forum?id=H1VGkIxRZ},
  eprint    = {1706.02690},
  archivePrefix = {arXiv}
}

@inproceedings{liu2020energy,
  author    = {Weitang Liu and Xiaoyun Wang and John Owens and Yixuan Li},
  title     = {Energy-based Out-of-distribution Detection},
  booktitle = {Advances in Neural Information Processing Systems},
  volume    = {33},
  pages     = {21464--21475},
  year      = {2020},
  eprint    = {2010.03759},
  archivePrefix = {arXiv}
}

@inproceedings{lee2018mahalanobis,
  author    = {Kimin Lee and Kibok Lee and Honglak Lee and Jinwoo Shin},
  title     = {A Simple Unified Framework for Detecting Out-of-Distribution Samples and Adversarial Attacks},
  booktitle = {Advances in Neural Information Processing Systems},
  volume    = {31},
  year      = {2018},
  x-note    = {the NeurIPS proceedings bibtex gives no pages},
  eprint    = {1807.03888},
  archivePrefix = {arXiv}
}

@inproceedings{sastry2020gram,
  author    = {Chandramouli Shama Sastry and Sageev Oore},
  title     = {Detecting Out-of-Distribution Examples with {G}ram Matrices},
  booktitle = {Proceedings of the 37th International Conference on Machine Learning},
  series    = {Proceedings of Machine Learning Research},
  volume    = {119},
  pages     = {8491--8501},
  year      = {2020},
  eprint    = {1912.12510},
  archivePrefix = {arXiv}
}

@inproceedings{sun2022knn,
  author    = {Yiyou Sun and Yifei Ming and Xiaojin Zhu and Yixuan Li},
  title     = {Out-of-Distribution Detection with Deep Nearest Neighbors},
  booktitle = {Proceedings of the 39th International Conference on Machine Learning},
  series    = {Proceedings of Machine Learning Research},
  volume    = {162},
  pages     = {20827--20840},
  year      = {2022},
  eprint    = {2204.06507},
  archivePrefix = {arXiv}
}

@inproceedings{wang2022vim,
  author    = {Haoqi Wang and Zhizhong Li and Litong Feng and Wayne Zhang},
  title     = {{ViM}: Out-of-Distribution With Virtual-Logit Matching},
  booktitle = {Proceedings of the IEEE/CVF Conference on Computer Vision and Pattern Recognition},
  pages     = {4921--4930},
  year      = {2022},
  eprint    = {2203.10807},
  archivePrefix = {arXiv}
}

@inproceedings{mueller2025mahapp,
  author    = {Maximilian M{\"u}ller and Matthias Hein},
  title     = {{M}ahalanobis++: Improving {OOD} Detection via Feature Normalization},
  booktitle = {Proceedings of the 42nd International Conference on Machine Learning},
  series    = {Proceedings of Machine Learning Research},
  volume    = {267},
  pages     = {45151--45184},
  year      = {2025},
  eprint    = {2505.18032},
  archivePrefix = {arXiv}
}

@misc{hossain2026mmpp,
  author    = {Rahim Hossain and Md Tawheedul Islam Bhuian and Md Farhan Shadiq and Kyoung-Don Kang},
  title     = {{MM++}: Post-Hoc Scale-Invariant Multilayer {OOD} Detection via Top-{K} Gated Feature Fusion},
  year      = {2026},
  note      = {arXiv preprint arXiv:2606.17352},
  x-note    = {title of v2; v1 said "Unsupervised" in place of "Post-Hoc"},
  eprint    = {2606.17352},
  archivePrefix = {arXiv}
}

@inproceedings{tack2020csi,
  author    = {Jihoon Tack and Sangwoo Mo and Jongheon Jeong and Jinwoo Shin},
  title     = {{CSI}: Novelty Detection via Contrastive Learning on Distributionally Shifted Instances},
  booktitle = {Advances in Neural Information Processing Systems},
  volume    = {33},
  pages     = {11839--11852},
  year      = {2020},
  eprint    = {2007.08176},
  archivePrefix = {arXiv}
}

@article{zhang2023openood15,
  author    = {Jingyang Zhang and Jingkang Yang and Pengyun Wang and Haoqi Wang and Yueqian Lin and Haoran Zhang and Yiyou Sun and Xuefeng Du and Yixuan Li and Ziwei Liu and Yiran Chen and Hai Li},
  title     = {{OpenOOD} v1.5: Enhanced Benchmark for Out-of-Distribution Detection},
  journal   = {Journal of Data-centric Machine Learning Research},
  year      = {2024},
  x-note    = {DMLR volume 2 (data.mlr.press/volumes/02); the official bib gives no volume},
  eprint    = {2306.09301},
  archivePrefix = {arXiv}
}

@inproceedings{du2022vos,
  author    = {Xuefeng Du and Zhaoning Wang and Mu Cai and Yixuan Li},
  title     = {{VOS}: Learning What You Don't Know by Virtual Outlier Synthesis},
  booktitle = {International Conference on Learning Representations},
  year      = {2022},
  eprint    = {2202.01197},
  archivePrefix = {arXiv}
}

@inproceedings{wilson2023safe,
  author    = {Samuel Wilson and Tobias Fischer and Feras Dayoub and Dimity Miller and Niko S{\"u}nderhauf},
  title     = {{SAFE}: Sensitivity-Aware Features for Out-of-Distribution Object Detection},
  booktitle = {Proceedings of the IEEE/CVF International Conference on Computer Vision},
  pages     = {23565--23576},
  year      = {2023},
  x-note    = {pages from CVF Open Access; IEEE Xplore (doi 10.1109/ICCV51070.2023.02154) gives 23508--23519}
}

@inproceedings{dhamija2020overlooked,
  author    = {Dhamija, Akshay Raj and G{\"u}nther, Manuel and Ventura, Jonathan and Boult, Terrance E.},
  title     = {The Overlooked Elephant of Object Detection: Open Set},
  booktitle = {Proceedings of the IEEE/CVF Winter Conference on Applications of Computer Vision},
  pages     = {1021--1030},
  year      = {2020},
  x-note    = {pages from the CVF Open Access PDF; IEEE Xplore (doi 10.1109/WACV45572.2020.9093355) gives 1010--1019}
}

@inproceedings{miller2018dropout,
  author    = {Miller, Dimity and Nicholson, Lachlan and Dayoub, Feras and S{\"u}nderhauf, Niko},
  title     = {Dropout Sampling for Robust Object Detection in Open-Set Conditions},
  booktitle = {2018 IEEE International Conference on Robotics and Automation},
  pages     = {3243--3249},
  year      = {2018},
  doi       = {10.1109/ICRA.2018.8460700}
}

@inproceedings{gupta2022owdetr,
  author    = {Gupta, Akshita and Narayan, Sanath and Joseph, K. J. and Khan, Salman and Khan, Fahad Shahbaz and Shah, Mubarak},
  title     = {{OW-DETR}: Open-World Detection Transformer},
  booktitle = {Proceedings of the IEEE/CVF Conference on Computer Vision and Pattern Recognition},
  pages     = {9235--9244},
  year      = {2022}
}

@inproceedings{du2022siren,
  author    = {Du, Xuefeng and Gozum, Gabriel and Ming, Yifei and Li, Yixuan},
  title     = {{SIREN}: Shaping Representations for Detecting Out-of-Distribution Objects},
  booktitle = {Advances in Neural Information Processing Systems},
  volume    = {35},
  pages     = {20434--20449},
  year      = {2022},
  doi       = {10.52202/068431-1486}
}

@inproceedings{kumar2023ffs,
  author    = {Kumar, Nishant and {\v{S}}egvi{\'c}, Sini{\v{s}}a and Eslami, Abouzar and Gumhold, Stefan},
  title     = {Normalizing Flow Based Feature Synthesis for Outlier-Aware Object Detection},
  booktitle = {Proceedings of the IEEE/CVF Conference on Computer Vision and Pattern Recognition},
  pages     = {5156--5165},
  year      = {2023},
  eprint    = {2302.07106},
  archivePrefix = {arXiv}
}

@inproceedings{wu2023srvae,
  author    = {Wu, Aming and Deng, Cheng},
  title     = {Discriminating Known from Unknown Objects via Structure-Enhanced Recurrent Variational AutoEncoder},
  booktitle = {Proceedings of the IEEE/CVF Conference on Computer Vision and Pattern Recognition},
  pages     = {23956--23965},
  year      = {2023}
}

@inproceedings{wu2023dfdd,
  author    = {Wu, Aming and Chen, Da and Deng, Cheng},
  title     = {Deep Feature Deblurring Diffusion for Detecting Out-of-Distribution Objects},
  booktitle = {Proceedings of the IEEE/CVF International Conference on Computer Vision},
  pages     = {13381--13391},
  year      = {2023}
}

@article{wu2024tib,
  author    = {Wu, Aming and Deng, Cheng},
  title     = {{TIB}: Detecting Unknown Objects via Two-Stream Information Bottleneck},
  journal   = {IEEE Transactions on Pattern Analysis and Machine Intelligence},
  volume    = {46},
  number    = {1},
  pages     = {611--625},
  year      = {2024},
  doi       = {10.1109/TPAMI.2023.3323523}
}

@inproceedings{wu2025wfs,
  author    = {Wu, Aming and Deng, Cheng},
  title     = {Percept, Memory, and Imagine: World Feature Simulating for Open-Domain Unknown Object Detection},
  booktitle = {Proceedings of the Computer Vision and Pattern Recognition Conference},
  pages     = {4682--4691},
  year      = {2025}
}

@inproceedings{liu2024syncood,
  author    = {Liu, Jiahui and Wen, Xin and Zhao, Shizhen and Chen, Yingxian and Qi, Xiaojuan},
  title     = {Can {OOD} Object Detectors Learn from Foundation Models?},
  booktitle = {European Conference on Computer Vision},
  pages     = {213--231},
  year      = {2024},
  doi       = {10.1007/978-3-031-73254-6_13},
  eprint    = {2409.05162},
  archivePrefix = {arXiv}
}

@inproceedings{zhang2025runa,
  author    = {Zhang, Bin and Chen, Jinggang and Qu, Xiaoyang and Li, Guokuan and Lu, Kai and Wan, Jiguang and Xiao, Jing and Wang, Jianzong},
  title     = {{RUNA}: Object-Level Out-of-Distribution Detection via Regional Uncertainty Alignment of Multimodal Representations},
  booktitle = {Proceedings of the AAAI Conference on Artificial Intelligence},
  volume    = {39},
  number    = {25},
  pages     = {26418--26426},
  year      = {2025},
  doi       = {10.1609/aaai.v39i25.34841},
  eprint    = {2503.22285},
  archivePrefix = {arXiv}
}

@inproceedings{li2026unioood,
  author    = {Li, Yuchuan and Motamedi, Azadeh and Kwon, Hyock Ju and Park, Chul B. and Kim, Il-Min},
  title     = {{UNI-OOD}: Unified Object- and Image-level Out-of-Distribution Detection via Cross-Context Attentive Vision-Language Modeling},
  booktitle = {Proceedings of the IEEE/CVF Conference on Computer Vision and Pattern Recognition},
  pages     = {6282--6292},
  year      = {2026}
}

@inproceedings{peng2026unoadapter,
  author    = {Peng, Ting and Dong, Junhao and Ong, Yew-Soon},
  title     = {Learning Latent Concepts for Detecting Out-of-Distribution Objects},
  booktitle = {Proceedings of the IEEE/CVF Conference on Computer Vision and Pattern Recognition},
  pages     = {28723--28733},
  year      = {2026},
  x-note    = {method name: UNO-Adapter},
}

@inproceedings{nguyen2026ronin,
  author    = {Nguyen, Quang-Huy and Zhou, Jin Peng and Liu, Zhenzhen and Bui, Khanh-Huyen and Weinberger, Kilian Q. and Chao, Wei-Lun and Le, Dung D.},
  title     = {Detecting Out-of-Distribution Objects through Class-Conditioned Inpainting},
  booktitle = {Proceedings of the IEEE/CVF Winter Conference on Applications of Computer Vision},
  pages     = {1937--1947},
  year      = {2026},
  eprint    = {2402.03292},
  archivePrefix = {arXiv},
  x-note    = {method name: RONIN},
}

@misc{gornhardt2026sam3o2d2,
  author    = {G{\"o}rnhardt, Lucas and Bartels, Timo and Fingscheidt, Tim},
  title     = {{SAM3-O2D2}: Zero-Shot Object Out-of-Distribution Detection by Object Class Prompting of the {SAM3}-Image Model},
  year      = {2026},
  note      = {arXiv preprint arXiv:2609.08281},
  eprint    = {2609.08281},
  archivePrefix = {arXiv}
}

@inproceedings{he2025yolohallucination,
  author    = {He, Weicheng and Wu, Changshun and Cheng, Chih-Hong and Huang, Xiaowei and Bensalem, Saddek},
  title     = {Mitigating Hallucinations in {YOLO}-based Object Detection Models: A Revisit to Out-of-Distribution Detection},
  booktitle = {2025 IEEE/RSJ International Conference on Intelligent Robots and Systems},
  pages     = {9870--9877},
  year      = {2025},
  doi       = {10.1109/IROS60139.2025.11245852}
}

@misc{montoya2025fmiyc,
  author    = {Montoya, Daniel and Bouguerra, Aymen and Gomez-Villa, Alexandra and Arnez, Fabio},
  title     = {{FindMeIfYouCan}: Bringing Open Set metrics to \textit{near}, \textit{far} and \textit{farther} Out-of-Distribution Object Detection},
  year      = {2025},
  note      = {arXiv preprint arXiv:2506.14008},
  x-note    = {ICLR 2026 submission, rejected (OpenReview decision 26 Jan 2026) -- cite arXiv only},
  eprint    = {2506.14008},
  archivePrefix = {arXiv}
}

@inproceedings{caldarella2025incremental,
  author    = {Caldarella, Simone and Ricci, Elisa and Aljundi, Rahaf},
  title     = {Incremental Object-Based Novelty Detection with Feedback Loop},
  booktitle = {International Conference on Image Analysis and Processing},
  pages     = {1--16},
  year      = {2025},
  doi       = {10.1007/978-3-032-10185-3_1},
  eprint    = {2311.09004},
  archivePrefix = {arXiv}
}

@inproceedings{torralba2011unbiased,
  author    = {Torralba, Antonio and Efros, Alexei A.},
  title     = {Unbiased Look at Dataset Bias},
  booktitle = {2011 IEEE Conference on Computer Vision and Pattern Recognition},
  pages     = {1521--1528},
  year      = {2011},
  doi       = {10.1109/CVPR.2011.5995347}
}

@inproceedings{liu2025decade,
  author    = {Liu, Zhuang and He, Kaiming},
  title     = {A Decade's Battle on Dataset Bias: Are We There Yet?},
  booktitle = {International Conference on Learning Representations},
  year      = {2025},
  eprint    = {2403.08632},
  archivePrefix = {arXiv}
}

@article{geirhos2020shortcut,
  author    = {Geirhos, Robert and Jacobsen, J{\"o}rn-Henrik and Michaelis, Claudio and Zemel, Richard and Brendel, Wieland and Bethge, Matthias and Wichmann, Felix A.},
  title     = {Shortcut Learning in Deep Neural Networks},
  journal   = {Nature Machine Intelligence},
  volume    = {2},
  number    = {11},
  pages     = {665--673},
  year      = {2020},
  doi       = {10.1038/s42256-020-00257-z},
  eprint    = {2004.07780},
  archivePrefix = {arXiv}
}

@inproceedings{singh2020context,
  author    = {Singh, Krishna Kumar and Mahajan, Dhruv and Grauman, Kristen and Lee, Yong Jae and Feiszli, Matt and Ghadiyaram, Deepti},
  title     = {Don't Judge an Object by Its Context: Learning to Overcome Contextual Bias},
  booktitle = {Proceedings of the IEEE/CVF Conference on Computer Vision and Pattern Recognition},
  pages     = {11070--11078},
  year      = {2020},
  eprint    = {2001.03152},
  archivePrefix = {arXiv}
}

@inproceedings{xiao2021noise,
  author    = {Xiao, Kai and Engstrom, Logan and Ilyas, Andrew and Madry, Aleksander},
  title     = {Noise or Signal: The Role of Image Backgrounds in Object Recognition},
  booktitle = {International Conference on Learning Representations},
  year      = {2021},
  eprint    = {2006.09994},
  archivePrefix = {arXiv}
}

@inproceedings{yang2024imagenetood,
  author    = {Yang, William and Zhang, Byron and Russakovsky, Olga},
  title     = {{ImageNet-OOD}: Deciphering Modern Out-of-Distribution Detection Algorithms},
  booktitle = {International Conference on Learning Representations},
  year      = {2024},
  eprint    = {2310.01755},
  archivePrefix = {arXiv}
}

@inproceedings{bitterwolf2023ninco,
  author    = {Bitterwolf, Julian and M{\"u}ller, Maximilian and Hein, Matthias},
  title     = {In or Out? Fixing {ImageNet} Out-of-Distribution Detection Evaluation},
  booktitle = {Proceedings of the 40th International Conference on Machine Learning},
  series    = {Proceedings of Machine Learning Research},
  volume    = {202},
  pages     = {2471--2506},
  year      = {2023},
  eprint    = {2306.00826},
  archivePrefix = {arXiv}
}

@inproceedings{ren2015faster,
  author    = {Ren, Shaoqing and He, Kaiming and Girshick, Ross and Sun, Jian},
  title     = {Faster {R-CNN}: Towards Real-Time Object Detection with Region Proposal Networks},
  booktitle = {Advances in Neural Information Processing Systems},
  volume    = {28},
  year      = {2015},
  x-note    = {the NeurIPS proceedings bibtex gives no pages}
}

@inproceedings{lin2017fpn,
  author    = {Lin, Tsung-Yi and Doll{\'a}r, Piotr and Girshick, Ross and He, Kaiming and Hariharan, Bharath and Belongie, Serge},
  title     = {Feature Pyramid Networks for Object Detection},
  booktitle = {Proceedings of the IEEE Conference on Computer Vision and Pattern Recognition},
  pages     = {2117--2125},
  year      = {2017}
}

@inproceedings{he2017maskrcnn,
  author    = {He, Kaiming and Gkioxari, Georgia and Doll{\'a}r, Piotr and Girshick, Ross},
  title     = {Mask {R-CNN}},
  booktitle = {Proceedings of the IEEE International Conference on Computer Vision},
  pages     = {2961--2969},
  year      = {2017}
}

@inproceedings{zhu2021deformable,
  author    = {Zhu, Xizhou and Su, Weijie and Lu, Lewei and Li, Bin and Wang, Xiaogang and Dai, Jifeng},
  title     = {Deformable {DETR}: Deformable Transformers for End-to-End Object Detection},
  booktitle = {International Conference on Learning Representations},
  year      = {2021}
}

@article{ledoit2004wellconditioned,
  author    = {Ledoit, Olivier and Wolf, Michael},
  title     = {A Well-Conditioned Estimator for Large-Dimensional Covariance Matrices},
  journal   = {Journal of Multivariate Analysis},
  volume    = {88},
  number    = {2},
  pages     = {365--411},
  year      = {2004},
  doi       = {10.1016/S0047-259X(03)00096-4}
}

@inproceedings{hsu2020godin,
  author    = {Yen-Chang Hsu and Yilin Shen and Hongxia Jin and Zsolt Kira},
  title     = {Generalized {ODIN}: Detecting Out-of-Distribution Image Without Learning From Out-of-Distribution Data},
  booktitle = {Proceedings of the IEEE/CVF Conference on Computer Vision and Pattern Recognition},
  pages     = {10951--10960},
  year      = {2020},
  x-note    = {cited as G-ODIN in VOS Table 1}
}

@inproceedings{lee2018training,
  author    = {Kimin Lee and Honglak Lee and Kibok Lee and Jinwoo Shin},
  title     = {Training Confidence-calibrated Classifiers for Detecting Out-of-Distribution Samples},
  booktitle = {International Conference on Learning Representations},
  year      = {2018},
  x-note    = {cited as GAN-synthesis in VOS Table 1}
}

@inproceedings{kuppers2020multivariate,
  author    = {Fabian K{\"u}ppers and Jan Kronenberger and Amirhossein Shantia and Anselm Haselhoff},
  title     = {Multivariate Confidence Calibration for Object Detection},
  booktitle = {Proceedings of the IEEE/CVF Conference on Computer Vision and Pattern Recognition Workshops},
  year      = {2020},
  x-note    = {neither the CVF bibtex nor the CVF PDF gives page numbers},
}

@inproceedings{gilg2023boxsize,
  author    = {Johannes Gilg and Torben Teepe and Fabian Herzog and Gerhard Rigoll},
  title     = {The Box Size Confidence Bias Harms Your Object Detector},
  booktitle = {Proceedings of the IEEE/CVF Winter Conference on Applications of Computer Vision},
  pages     = {1471--1480},
  year      = {2023}
}

@inproceedings{wu2020rethinking,
  author    = {Yue Wu and Yinpeng Chen and Lu Yuan and Zicheng Liu and Lijuan Wang and Hongzhi Li and Yun Fu},
  title     = {Rethinking Classification and Localization for Object Detection},
  booktitle = {Proceedings of the IEEE/CVF Conference on Computer Vision and Pattern Recognition},
  pages     = {10186--10195},
  year      = {2020}
}

@article{bates2023testing,
  author    = {Stephen Bates and Emmanuel Cand{\`e}s and Lihua Lei and Yaniv Romano and Matteo Sesia},
  title     = {Testing for Outliers with Conformal p-values},
  journal   = {The Annals of Statistics},
  volume    = {51},
  number    = {1},
  pages     = {149--178},
  year      = {2023},
  doi       = {10.1214/22-AOS2244}
}

@inproceedings{harun2025controlling,
  author    = {Md Yousuf Harun and Jhair Gallardo and Christopher Kanan},
  title     = {Controlling Neural Collapse Enhances Out-of-Distribution Detection and Transfer Learning},
  booktitle = {Proceedings of the 42nd International Conference on Machine Learning},
  series    = {Proceedings of Machine Learning Research},
  volume    = {267},
  pages     = {22149--22176},
  year      = {2025}
}

@inproceedings{liu2020simple,
  author    = {Jeremiah Liu and Zi Lin and Shreyas Padhy and Dustin Tran and Tania {Bedrax Weiss} and Balaji Lakshminarayanan},
  title     = {Simple and Principled Uncertainty Estimation with Deterministic Deep Learning via Distance Awareness},
  booktitle = {Advances in Neural Information Processing Systems},
  volume    = {33},
  pages     = {7498--7512},
  year      = {2020},
  eprint    = {2006.10108},
  archivePrefix = {arXiv}
}

@inproceedings{vanamersfoort2020uncertainty,
  author    = {Van Amersfoort, Joost and Smith, Lewis and Teh, Yee Whye and Gal, Yarin},
  title     = {Uncertainty Estimation Using a Single Deep Deterministic Neural Network},
  booktitle = {Proceedings of the 37th International Conference on Machine Learning},
  series    = {Proceedings of Machine Learning Research},
  volume    = {119},
  pages     = {9690--9700},
  year      = {2020},
  eprint    = {2003.02037},
  archivePrefix = {arXiv}
}

@misc{postels2020hidden,
  author    = {Janis Postels and Hermann Blum and Yannick Str{\"u}mpler and Cesar Cadena and Roland Siegwart and Luc Van Gool and Federico Tombari},
  title     = {The Hidden Uncertainty in a Neural Network's Activations},
  year      = {2020},
  note      = {arXiv preprint arXiv:2012.03082},
  x-note    = {arXiv v2 title; the arXiv metadata drops the apostrophe, the PDF keeps it. Also at the ICML 2021 UDL workshop (non-archival)},
  eprint    = {2012.03082},
  archivePrefix = {arXiv}
}

@misc{vovk2003mondrian,
  author = {Vladimir Vovk and David Lindsay and Ilia Nouretdinov and Alex Gammerman},
  title  = {Mondrian Confidence Machine},
  note   = {On-line Compression Modelling Project, Working Paper~\#4, \url{https://alrw.cs.rhul.ac.uk/old/04.pdf}},
  year   = {2003},
}

@misc{angelopoulos2024theoretical,
  author  = {Anastasios N. Angelopoulos and Rina Foygel Barber and Stephen Bates},
  title   = {Theoretical Foundations of Conformal Prediction},
  year    = {2024},
  note    = {arXiv preprint arXiv:2411.11824},
  x-note  = {arXiv v5 states it will be published by Cambridge University Press},
  eprint  = {2411.11824},
  archivePrefix = {arXiv}
}

@misc{macedo2022dismax,
  author    = {David Mac{\^e}do and Cleber Zanchettin and Teresa Ludermir},
  title     = {Distinction Maximization Loss: Efficiently Improving Out-of-Distribution Detection and Uncertainty Estimation by Replacing the Loss and Calibrating},
  year      = {2022},
  note      = {arXiv preprint arXiv:2205.05874},
  x-note    = {arXiv v5 title; earlier versions had other titles},
  eprint    = {2205.05874},
  archivePrefix = {arXiv}
}

@inproceedings{haroush2022statistical,
  author    = {Matan Haroush and Tzviel Frostig and Ruth Heller and Daniel Soudry},
  title     = {A Statistical Framework for Efficient Out of Distribution Detection in Deep Neural Networks},
  booktitle = {International Conference on Learning Representations},
  year      = {2022},
  eprint    = {2102.12967},
  archivePrefix = {arXiv}
}

@inproceedings{timans2024adaptive,
  author    = {Alexander Timans and Christoph-Nikolas Straehle and Kaspar Sakmann and Eric Nalisnick},
  title     = {Adaptive Bounding Box Uncertainties via Two-Step Conformal Prediction},
  booktitle = {European Conference on Computer Vision},
  pages     = {363--398},
  year      = {2024},
  doi       = {10.1007/978-3-031-73223-2_21},
  eprint    = {2403.07263},
  archivePrefix = {arXiv}
}

@inproceedings{ma2025online,
  author    = {Xinsong Ma and Xin Zou and Weiwei Liu},
  title     = {An Online Statistical Framework for Out-of-Distribution Detection},
  booktitle = {Proceedings of the 42nd International Conference on Machine Learning},
  series    = {Proceedings of Machine Learning Research},
  volume    = {267},
  pages     = {42309--42324},
  year      = {2025}
}
}

\clearpage
\setcounter{section}{0}
\setcounter{figure}{0}
\setcounter{table}{0}
\setcounter{equation}{0}

\section{Implementation details}
\label{sec:s_impl}

All experiments run on one GPU with the software of Table~\ref{tab:s_env}.
The detectors are the public ResNet-50 checkpoints of VOS~\cite{du2022vos} for Faster R-CNN and of SIREN~\cite{du2022siren} for Deformable DETR, and PLUM leaves their weights unchanged.
As in VOS, Faster R-CNN keeps a detection when its largest foreground softmax reaches the threshold that maximizes the F1 score on the ID validation set.
The only random step of PLUM draws the BDD100K training images with seed 0.
We disable TF32 arithmetic on the GPU.

\begin{table}[t]
  \centering
  \caption{Hardware, software and settings of the experiments.}
  \label{tab:s_env}
  \footnotesize
  \begin{tabular}{ll}
    \toprule
    Item & Setting \\
    \midrule
    GPU & NVIDIA GeForce RTX 4090 \\
    Python & 3.12 \\
    PyTorch & 2.14 with CUDA 13.2 \\
    detectron2 & 0.6, built from source \\
    Faster R-CNN & ResNet-50 FPN, VOS checkpoints \\
    Deformable DETR & ResNet-50, SIREN checkpoints \\
    Detection threshold & 0.5728 PASCAL-VOC, 0.4006 BDD100K \\
    Training images & 16{,}551 PASCAL-VOC, 10{,}000 BDD100K \\
    Box features & half precision \\
    Statistics and distances & double precision \\
    \bottomrule
  \end{tabular}
\end{table}

\section{Deformable DETR}
\label{sec:s_ddetr}

We use the public Deformable DETR~\cite{zhu2021deformable} checkpoints with a ResNet-50 backbone released with SIREN~\cite{du2022siren} and follow its protocol, which evaluates every pair of query and class with a sigmoid output above 0.1 in the last decoder layer.
PLUM reads the last backbone stage inside the output box, because the detector has no proposals.
The size level follows the same feature-pyramid rule as in Faster R-CNN, and the ID statistics, the lines and the references come from the training images as in Sec.~\ref{sec:method_level} of the main paper.
PLUM keeps the detections unchanged, so the detector keeps its AP of 60.5 on PASCAL-VOC, averaged over IoU thresholds from 0.5 to 0.95.
As in SIREN, we leave out MSP and the energy score because the head outputs sigmoids.
KNN and SIREN-KNN, which reads the projection features of the SIREN checkpoint, use the number of neighbors and the reference bank that are best on each OOD set, which favors them.

\begin{table*}[t]
  \centering
  \caption{OOD object detection with a ResNet-50 Deformable DETR. Bold marks the best value in each column. $^\dagger$Published MLP. $^\ddagger$Reported in~\cite{du2022siren}.}
  \label{tab:s_ddetr}
  \footnotesize
  \begin{tabular}{l c cccc cccc}
    \toprule
     & & \multicolumn{4}{c}{\textbf{ID: PASCAL-VOC}} & \multicolumn{4}{c}{\textbf{ID: BDD100K}} \\
    \cmidrule(lr){3-6} \cmidrule(lr){7-10}
     & & \multicolumn{2}{c}{MS-COCO} & \multicolumn{2}{c}{OpenImages} & \multicolumn{2}{c}{MS-COCO} & \multicolumn{2}{c}{OpenImages} \\
    \cmidrule(lr){3-4} \cmidrule(lr){5-6} \cmidrule(lr){7-8} \cmidrule(lr){9-10}
    Method & Retrain? & FPR95$\downarrow$ & AUROC$\uparrow$ & FPR95$\downarrow$ & AUROC$\uparrow$ & FPR95$\downarrow$ & AUROC$\uparrow$ & FPR95$\downarrow$ & AUROC$\uparrow$ \\
    \midrule
    Mahalanobis~\cite{lee2018mahalanobis} &  & 97.36 & 50.44 & 97.84 & 49.12 & 70.97 & 79.89 & 67.48 & 83.42 \\
    KNN~\cite{sun2022knn} &  & 94.66 & 59.51 & 94.53 & 57.05 & 54.08 & 87.61 & 46.90 & 91.58 \\
    Gram~\cite{sastry2020gram}$^\ddagger$ &  & 94.16 & 43.97 & 95.29 & 38.81 & 73.81 & 60.13 & 71.56 & 57.14 \\
    SAFE~\cite{wilson2023safe}$^\dagger$ &  & 50.38 & 77.25 & \textbf{7.32} & \textbf{97.48} & 39.35 & 85.55 & 21.81 & 94.30 \\
    \midrule
    CSI~\cite{tack2020csi}$^\ddagger$ & \checkmark & 84.00 & 55.07 & 79.16 & 51.37 & 70.27 & 77.93 & 71.30 & 76.42 \\
    VOS~\cite{du2022vos}$^\ddagger$ & \checkmark & 97.46 & 54.40 & 97.07 & 52.77 & 76.44 & 77.33 & 72.58 & 76.62 \\
    OW-DETR~\cite{gupta2022owdetr}$^\ddagger$ & \checkmark & 93.09 & 55.70 & 93.82 & 57.80 & 80.78 & 70.29 & 77.37 & 73.78 \\
    Dismax~\cite{macedo2022dismax}$^\ddagger$ & \checkmark & 82.05 & 75.21 & 76.37 & 70.66 & 77.62 & 72.14 & 81.23 & 67.18 \\
    SIREN-vMF~\cite{du2022siren}$^\ddagger$ & \checkmark & 75.49 & 76.10 & 78.36 & 71.05 & 67.54 & 80.06 & 66.31 & 79.77 \\
    SIREN-KNN~\cite{du2022siren} & \checkmark & 64.03 & 78.95 & 67.74 & 75.37 & 54.77 & 86.06 & 48.15 & 89.24 \\
    \midrule
    \textbf{PLUM (Ours)} &  & \textbf{43.17} & \textbf{90.74} & 25.59 & 95.07 & \textbf{37.34} & \textbf{93.00} & \textbf{19.29} & \textbf{95.78} \\
    \bottomrule
  \end{tabular}
\end{table*}

Table~\ref{tab:s_ddetr} reports the benchmark.
PLUM has the highest AUROC and the lowest FPR95 in three of the four pairings.
With OpenImages as OOD on PASCAL-VOC, SAFE reaches 97.48 against 95.07 for PLUM.

\begin{table}[t]
  \centering
  \caption{Size and power of the backbone distance of Deformable DETR under different null distributions, averaged over MS-COCO and OpenImages as OOD sets. False alarms give the range over strata of the share of ID boxes with rank at most 5\%.}
  \label{tab:s_ddetr_null}
  \footnotesize
  \setlength{\tabcolsep}{2.5pt}
  \resizebox{\linewidth}{!}{%
  \begin{tabular}{l ccc ccc}
    \toprule
     & \multicolumn{3}{c}{PASCAL-VOC} & \multicolumn{3}{c}{BDD100K} \\
    \cmidrule(lr){2-4} \cmidrule(lr){5-7}
    Null distribution & AUROC$\uparrow$ & FPR95$\downarrow$ & False alarms & AUROC$\uparrow$ & FPR95$\downarrow$ & False alarms \\
    \midrule
    One for all boxes & 80.5 & 75.4 & 0.2--15.3 & 95.8 & 24.6 & 2.0--19.7 \\
    Size level & 92.4 & 38.0 & 3.4--6.6 & \textbf{95.9} & \textbf{20.5} & 3.9--19.6 \\
    Size level $\times$ border & 92.5 & 38.0 & 3.7--4.8 & 94.8 & 25.5 & 3.4--7.9 \\
    PLUM, with line & \textbf{92.9} & \textbf{34.4} & 3.5--4.8 & 94.4 & 28.3 & 2.6--6.3 \\
    \bottomrule
  \end{tabular}%
  }
\end{table}

\begin{figure}[t]
  \centering
  \includegraphics[width=\linewidth]{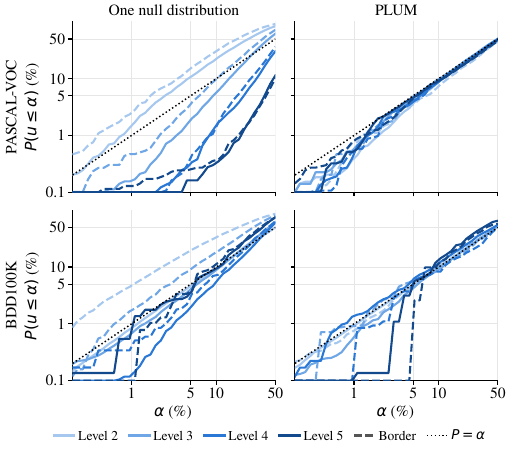}
  \caption{Share of ID validation boxes with rank at most $\alpha$ in each stratum, with a ResNet-50 Deformable DETR. A valid test follows the dotted diagonal. Line color gives the size level of the stratum, and dashed lines mark boxes at the image border. Shares below 0.1\% are drawn at 0.1\%.}
  \label{fig:s_ddetr_calib}
\end{figure}

Table~\ref{tab:s_ddetr_null} and Fig.~\ref{fig:s_ddetr_calib} repeat the size analysis of Sec.~\ref{sec:exp_size} of the main paper.
As with Faster R-CNN, one null distribution spreads the strata around the diagonal, and PLUM brings them onto it.
At $\alpha = 5\%$, the strata range from 0.2\% to 15.3\% on PASCAL-VOC and from 2.0\% to 19.7\% on BDD100K with one null distribution, and from 3.5\% to 4.8\% and from 2.6\% to 6.3\% with PLUM.
On BDD100K, border contact is again needed to hold the level, since size alone leaves a stratum at 19.6\%.
With this detector, the conditioning on BDD100K lowers the AUROC from 95.8 with one null distribution to 94.4 with PLUM.
The loss comes from border contact, as the step from the size level to the size level and border contact in Table~\ref{tab:s_ddetr_null} shows.
On BDD100K, 2.7\% of the ID detections touch the image border against 31.7\% and 33.8\% of the OOD detections with MS-COCO and OpenImages as OOD, so border contact alone separates them with an AUROC of 64.5 and 65.5.
A box at the border also has a large distance whether it is ID or OOD.
Ranked against all boxes, 16.8\% of the ID boxes at the border fall below the 5\% level against 4.6\% of the others, so one null distribution credits border contact as evidence.
Ranking within strata removes this credit.
From the size level to the size level and border contact, the AUROC of the OOD boxes at the border against all ID boxes falls from 95.3 to 90.3 with MS-COCO as OOD and from 97.2 to 94.0 with OpenImages, while that of the other OOD boxes moves from 94.4 to 94.7 and from 97.1 to 97.3.
The size level itself adds only 0.1 to the AUROC of this detector on BDD100K, so nothing offsets this loss.
On PASCAL-VOC, where 36.9\% of the ID detections touch the border, conditioning raises the AUROC from 80.5 to 92.9.

\section{Layers and false alarms}
\label{sec:s_details}

\begin{table*}[t]
  \centering
  \caption{The PLUM distance read at different layers for each OOD set. Raw uses one null distribution and Leveled applies Sec.~\ref{sec:method_level} of the main paper. Rules mark the task boundary, and bold marks the best Leveled value of each detector. BDD100K has no feature-pyramid dump.}
  \label{tab:s_layers}
  \footnotesize
  \setlength{\tabcolsep}{2.2pt}
  \resizebox{\textwidth}{!}{%
  \begin{tabular}{l cccc cccc cccc cccc}
    \toprule
     & \multicolumn{8}{c}{\textbf{ID: PASCAL-VOC}} & \multicolumn{8}{c}{\textbf{ID: BDD100K}} \\
    \cmidrule(lr){2-9} \cmidrule(lr){10-17}
     & \multicolumn{4}{c}{MS-COCO} & \multicolumn{4}{c}{OpenImages} & \multicolumn{4}{c}{MS-COCO} & \multicolumn{4}{c}{OpenImages} \\
    \cmidrule(lr){2-5} \cmidrule(lr){6-9} \cmidrule(lr){10-13} \cmidrule(lr){14-17}
     & \multicolumn{2}{c}{AUROC$\uparrow$} & \multicolumn{2}{c}{FPR95$\downarrow$} & \multicolumn{2}{c}{AUROC$\uparrow$} & \multicolumn{2}{c}{FPR95$\downarrow$} & \multicolumn{2}{c}{AUROC$\uparrow$} & \multicolumn{2}{c}{FPR95$\downarrow$} & \multicolumn{2}{c}{AUROC$\uparrow$} & \multicolumn{2}{c}{FPR95$\downarrow$} \\
    \cmidrule(lr){2-3} \cmidrule(lr){4-5} \cmidrule(lr){6-7} \cmidrule(lr){8-9} \cmidrule(lr){10-11} \cmidrule(lr){12-13} \cmidrule(lr){14-15} \cmidrule(lr){16-17}
    Layer & Raw & Leveled & Raw & Leveled & Raw & Leveled & Raw & Leveled & Raw & Leveled & Raw & Leveled & Raw & Leveled & Raw & Leveled \\
    \midrule
    \multicolumn{17}{l}{\textit{Faster R-CNN}} \\
    res2 & 66.3 & 80.4 & 88.9 & 65.2 & 78.7 & 93.3 & 76.3 & 33.3 & 52.9 & 87.4 & 92.4 & 57.2 & 54.1 & 91.8 & 92.6 & 46.0 \\
    res3 & 61.8 & 82.8 & 93.8 & 65.0 & 72.0 & 94.0 & 93.5 & 29.6 & 50.4 & 90.3 & 93.7 & 47.8 & 53.3 & 94.4 & 93.7 & 30.0 \\
    res4 & 61.1 & 86.1 & 91.4 & 58.3 & 65.9 & 93.4 & 92.3 & 33.0 & 74.9 & 93.1 & 79.0 & 40.5 & 81.4 & 96.6 & 70.9 & 19.4 \\
    res5 & 74.0 & \textbf{91.5} & 79.0 & \textbf{44.2} & 74.4 & \textbf{94.5} & 81.3 & \textbf{29.2} & 93.5 & \textbf{95.1} & 39.6 & \textbf{24.3} & 95.2 & \textbf{97.5} & 38.0 & \textbf{13.9} \\
    \midrule
    FPN & 57.6 & 82.4 & 89.6 & 66.8 & 50.3 & 82.1 & 94.9 & 63.2 & -- & -- & -- & -- & -- & -- & -- & -- \\
    fc1 & 56.3 & 60.7 & 97.8 & 97.2 & 53.1 & 59.7 & 98.2 & 96.2 & 83.8 & 87.4 & 78.9 & 64.7 & 85.0 & 89.0 & 77.1 & 59.3 \\
    fc2 & 59.4 & 64.8 & 96.8 & 95.6 & 57.6 & 65.3 & 96.3 & 93.8 & 87.4 & 86.5 & 59.5 & 68.8 & 88.2 & 86.7 & 61.0 & 71.6 \\
    logit & 78.1 & 77.1 & 70.6 & 74.4 & 78.0 & 76.8 & 66.5 & 72.3 & 85.8 & 83.2 & 53.4 & 58.4 & 87.8 & 85.1 & 46.4 & 51.9 \\
    \midrule
    \multicolumn{17}{l}{\textit{Deformable DETR}} \\
    backbone & 80.1 & \textbf{90.7} & 76.3 & \textbf{43.2} & 80.8 & \textbf{95.1} & 74.4 & \textbf{25.6} & 94.4 & \textbf{93.0} & 34.1 & \textbf{37.3} & 97.3 & \textbf{95.8} & 15.1 & \textbf{19.3} \\
    \midrule
    decoder & 56.2 & 50.1 & 94.3 & 96.7 & 59.0 & 47.5 & 93.8 & 98.0 & 93.0 & 87.3 & 31.8 & 61.1 & 95.6 & 87.9 & 23.7 & 60.7 \\
    \bottomrule
  \end{tabular}%
  }
\end{table*}

\begin{table*}[t]
  \centering
  \caption{Share of ID validation boxes with rank at most $\alpha$ under PLUM in each stratum, with a ResNet-50 Faster R-CNN, in percent.}
  \label{tab:s_alpha}
  \footnotesize
  \setlength{\tabcolsep}{3pt}
  \begin{tabular}{l l ccccc ccccc}
    \toprule
     & & \multicolumn{5}{c}{\textbf{ID: PASCAL-VOC}} & \multicolumn{5}{c}{\textbf{ID: BDD100K}} \\
    \cmidrule(lr){3-7} \cmidrule(lr){8-12}
    Level & Contact & ID & Reference & 1\% & 5\% & 10\% & ID & Reference & 1\% & 5\% & 10\% \\
    \midrule
    2 & Inner & 3{,}210 & 2{,}585 & 1.09 & 5.58 & 11.59 & 180{,}169 & 44{,}333 & 0.95 & 5.08 & 10.11 \\
    2 & Border & 241 & 217 & 0.41 & 0.83 & 2.90 & 4{,}031 & 1{,}065 & 0.55 & 4.07 & 9.15 \\
    3 & Inner & 3{,}299 & 2{,}424 & 0.61 & 5.24 & 11.28 & 12{,}972 & 3{,}414 & 0.78 & 4.39 & 9.71 \\
    3 & Border & 544 & 362 & 1.47 & 7.54 & 10.11 & 2{,}039 & 561 & 1.28 & 4.32 & 11.23 \\
    4 & Inner & 3{,}581 & 2{,}813 & 0.92 & 4.41 & 8.91 & 4{,}511 & 1{,}136 & 1.35 & 4.66 & 10.71 \\
    4 & Border & 1{,}348 & 975 & 0.37 & 5.34 & 9.05 & 1{,}998 & 430 & 0.15 & 2.60 & 7.11 \\
    5 & Inner & 2{,}089 & 1{,}695 & 0.34 & 3.69 & 10.00 & 258 & 68 & 0.00 & 4.65 & 6.98 \\
    5 & Border & 2{,}843 & 2{,}382 & 0.46 & 4.89 & 9.53 & 361 & 97 & 0.00 & 5.82 & 10.80 \\
    \midrule
    \multicolumn{2}{l}{All boxes} & 17{,}155 & -- & 0.71 & 4.90 & 10.07 & 206{,}339 & -- & 0.93 & 4.98 & 10.06 \\
    \bottomrule
  \end{tabular}
\end{table*}

Table~\ref{tab:s_layers} gives the layer comparison of Table~\ref{tab:layers} of the main paper for each OOD set.
In every pairing, res5 has the best Leveled AUROC and FPR95 among the layers of Faster R-CNN, and the backbone of Deformable DETR has a better Leveled AUROC and FPR95 than its decoder.

Table~\ref{tab:s_alpha} gives the values behind Fig.~\ref{fig:calib} of the main paper.
Over all boxes the share stays within 0.3 points of $\alpha$ at every level.

\begin{table}[t]
  \centering
  \caption{Strata whose share of ID validation boxes with rank at most $\alpha$ lies below or above its 95\% range under exchangeable references.}
  \label{tab:s_exch}
  \footnotesize
  \setlength{\tabcolsep}{3pt}
  \resizebox{\linewidth}{!}{%
  \begin{tabular}{l l c cc cc cc}
    \toprule
     & & & \multicolumn{2}{c}{$\alpha = 1\%$} & \multicolumn{2}{c}{$\alpha = 5\%$} & \multicolumn{2}{c}{$\alpha = 10\%$} \\
    \cmidrule(lr){4-5} \cmidrule(lr){6-7} \cmidrule(lr){8-9}
    Detector & ID & Strata & Below & Above & Below & Above & Below & Above \\
    \midrule
    Faster R-CNN & PASCAL-VOC & 8 & 2 & 0 & 1 & 0 & 1 & 0 \\
     & BDD100K & 8 & 1 & 0 & 1 & 0 & 1 & 0 \\
    \midrule
    Deformable DETR & PASCAL-VOC & 8 & 5 & 0 & 4 & 0 & 5 & 0 \\
     & BDD100K & 7 & 0 & 0 & 1 & 0 & 1 & 0 \\
    \bottomrule
  \end{tabular}%
  }
\end{table}

The reference boxes of PLUM come from the images on which the detector was trained, whereas the ID validation boxes come from images that it has not seen.
Table~\ref{tab:s_exch} compares the share in each stratum with the range that it would have if the two kinds of boxes were exchangeable.
Let a stratum have $n$ reference boxes and let $l = \lfloor (n + 1)\alpha \rfloor$.
Given the references, the false-alarm rate of the stratum then follows a $\mathrm{Beta}(l, n + 1 - l)$ distribution~\cite{angelopoulos2024theoretical}, as Sec.~\ref{sec:s_derivations} derives.
The share among the $n_{\mathrm{ID}}$ ID boxes of the stratum follows the corresponding beta-binomial distribution, and we take its central 95\% range.
Like the references, the comparison counts a proposal or a query that the detector outputs with several classes only once.

No stratum lies above its range at any level in either detector.

The strata that leave the range all lie below it, in 23 of the 93 pairs of stratum and level.
Boxes at the border at level 3 on PASCAL-VOC flag 7.54\% at $\alpha = 5\%$ and lie inside their range from 2.39\% to 8.09\%.
The two strata of the largest boxes on BDD100K have 68 and 97 reference boxes, so $l = 0$ at $\alpha = 1\%$ and their share is zero for any score.
Zero is also the only share that exchangeable references allow there.
Under exact exchangeability, about 5\% of the pairs would still leave a 95\% range.
The range also treats the boxes of one image as independent, which makes it narrower than the spread under the dependence within images.

\begin{table}[t]
  \centering
  \caption{Largest $\delta$ over strata in points for every null distribution of Table~\ref{tab:null} of the main paper. The last column gives the 95th percentile of the largest $\delta$ of PLUM over resampled ID validation images with the references fixed.}
  \label{tab:s_delta}
  \footnotesize
  \setlength{\tabcolsep}{3pt}
  \resizebox{\linewidth}{!}{%
  \begin{tabular}{l l cccc c}
    \toprule
    Detector & ID & One null & Size level & Size level $\times$ border & PLUM & Resampled \\
    \midrule
    Faster R-CNN & PASCAL-VOC & 27.0 & 1.8 & 3.4 & 3.0 & 5.2 \\
     & BDD100K & 49.1 & 50.7 & 5.5 & 5.2 & 8.5 \\
    \midrule
    Deformable DETR & PASCAL-VOC & 21.0 & 3.2 & 0.0 & 0.1 & 1.4 \\
     & BDD100K & 22.7 & 22.7 & 8.2 & 2.1 & 4.4 \\
    \bottomrule
  \end{tabular}%
  }
\end{table}

Table~\ref{tab:s_delta} extends Table~\ref{tab:exch} of the main paper to every null distribution of Table~\ref{tab:null} of the main paper, and Sec.~\ref{sec:s_derivations} derives the bound.
On BDD100K with Faster R-CNN, the size level alone leaves small boxes at the border at 50.7 points, and border contact brings the largest $\delta$ to 5.5.
On BDD100K with Deformable DETR, the line in log area lowers the largest $\delta$ from 8.2 to 2.1.
The two strata where $\delta$ of PLUM is significant hold small boxes away from the border, with 1.75 points on PASCAL-VOC with Faster R-CNN and 0.19 points on BDD100K with Deformable DETR.
Under exact exchangeability about 1.6 of the 31 strata would be significant.
The significance treats the boxes of one image as independent, so it flags a shift more readily than the dependence within images warrants.
Under exact exchangeability, about 5\% of the pairs would still leave a 95\% range.
The range also treats the boxes of one image as independent, which makes it narrower than the spread under the dependence within images.
\clearpage
\clearpage
\section{Checks of the compared scores}
\label{sec:s_checks}

\begin{figure}[t]
  \centering
  \includegraphics[width=\linewidth]{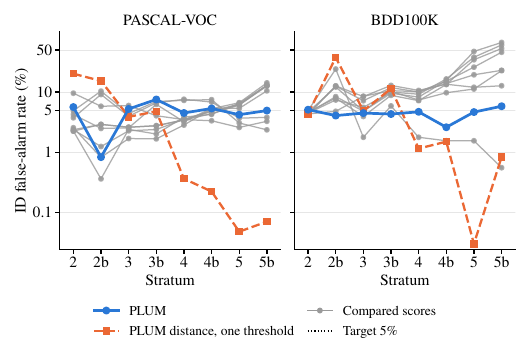}
  \caption{ID false-alarm rate in each stratum at one threshold that keeps 95\% of the ID detections, with a ResNet-50 Faster R-CNN. The number gives the feature-pyramid level of the stratum, and b marks boxes that touch the image border. Rates below 0.03\% are drawn at 0.03\%. The gray lines are the compared scores of Table~\ref{tab:s_checks}.}
  \label{fig:s_strata}
\end{figure}

Table~\ref{tab:s_checks} and Fig.~\ref{fig:s_strata} apply the checks of Sec.~\ref{sec:formulation} of the main paper to every score that we evaluate ourselves on the ResNet-50 Faster R-CNN.
The background AUROC $A_{\mathrm{bg}}$ and the perturbation of the boxes follow Sec.~\ref{sec:exp_object} of the main paper, and we report the departure $|A_{\mathrm{bg}} - 50|$ and the AUROC drop $\Delta_{\mathrm{jit}}$ at $\varepsilon = 0.5$.
Two further quantities enter the table.
The source check compares PASCAL-VOC detections that match a ground-truth object of the predicted class at IoU above 0.3 with MS-COCO detections that match such an object of a PASCAL-VOC class.
$\Delta_{\mathrm{src}}$ is the change in AUROC against the MS-COCO OOD set when the ID detections come from the MS-COCO images, and it stays near zero for a score that does not read the source of the image.
$R_{\max}$ is the largest ID false-alarm rate over the strata with at least 200 ID detections at one threshold that keeps 95\% of the ID detections, and it stays near 5\% when the false alarms spread evenly.

\begin{table*}[t]
  \centering
  \caption{Checks of the compared scores with a ResNet-50 Faster R-CNN. $\Delta_{\mathrm{src}}$ uses MS-COCO as OOD and should stay near zero. $R_{\max}$ uses ID data only and is nominally 5. SIREN has no checkpoint for BDD100K.}
  \label{tab:s_checks}
  \footnotesize
  \setlength{\tabcolsep}{4pt}
  \begin{tabular}{l cccc c cc cccc}
    \toprule
     & \multicolumn{5}{c}{Image} & \multicolumn{2}{c}{Geometry} & \multicolumn{4}{c}{Placement} \\
    \cmidrule(lr){2-6} \cmidrule(lr){7-8} \cmidrule(lr){9-12}
     & \multicolumn{4}{c}{$|A_{\mathrm{bg}} - 50|\downarrow$} & $\Delta_{\mathrm{src}}$ & \multicolumn{2}{c}{$R_{\max}\downarrow$} & \multicolumn{4}{c}{$\Delta_{\mathrm{jit}}\downarrow$} \\
    \cmidrule(lr){2-5} \cmidrule(lr){6-6} \cmidrule(lr){7-8} \cmidrule(lr){9-12}
     & \multicolumn{2}{c}{VOC} & \multicolumn{2}{c}{BDD} & VOC & VOC & BDD & \multicolumn{2}{c}{VOC} & \multicolumn{2}{c}{BDD} \\
    \cmidrule(lr){2-3} \cmidrule(lr){4-5} \cmidrule(lr){9-10} \cmidrule(lr){11-12}
    Method & COCO & OI & COCO & OI & COCO & & & COCO & OI & COCO & OI \\
    \midrule
    MSP & 2.2 & 0.3 & 6.4 & 5.1 & 2.1 & 12.9 & 46.0 & 1.1 & 1.2 & $-$0.6 & $-$0.7 \\
    ODIN & 5.7 & 2.8 & 10.7 & 11.3 & $-$0.9 & 13.0 & 60.7 & 1.2 & 1.1 & $-$1.4 & $-$1.6 \\
    Energy & 5.2 & 0.1 & 13.7 & 15.4 & $-$1.2 & 12.6 & 52.9 & 1.6 & 1.3 & $-$0.3 & $-$0.7 \\
    Mahalanobis & 4.4 & 17.4 & 14.4 & 26.7 & 3.4 & 10.4 & 13.5 & 10.2 & 8.4 & 15.4 & 15.7 \\
    KNN & 8.6 & 0.0 & 7.9 & 4.2 & $-$2.0 & 10.5 & 22.7 & 2.6 & 2.9 & 5.5 & 4.6 \\
    ViM & 7.3 & 17.6 & 12.7 & 25.2 & 0.1 & 9.1 & 23.0 & 12.4 & 11.6 & 11.7 & 11.9 \\
    SAFE & 20.7 & 39.3 & 43.3 & 43.5 & $-$28.4 & 9.6 & 24.1 & 0.1 & 0.1 & $-$0.2 & $-$0.1 \\
    VOS & 8.9 & 0.8 & 12.8 & 13.1 & 0.1 & 14.2 & 66.8 & 1.4 & 1.8 & 0.4 & 0.4 \\
    SIREN & 1.9 & 9.4 & -- & -- & 1.5 & 12.9 & -- & 2.0 & 2.1 & -- & -- \\
    \midrule
    \textbf{PLUM (Ours)} & 5.3 & 23.5 & 3.7 & 5.9 & $-$0.6 & 7.5 & 5.8 & 0.6 & 0.4 & 0.2 & 0.3 \\
    \bottomrule
  \end{tabular}
\end{table*}

On the image, SAFE departs from 50 by up to 44 points on background boxes, and its AUROC falls by 28.4 points when the ID detections come from MS-COCO images.
PLUM departs by 23.5 points with OpenImages as OOD on PASCAL-VOC and by at most 6 points elsewhere, and its AUROC changes by 0.6 points under the source check.
On the geometry, the worst stratum of the compared scores reaches 9.1\% to 14.2\% on PASCAL-VOC and 13.5\% to 66.8\% on BDD100K, against 7.5\% and 5.8\% for PLUM.
On the placement, the scores that read fc2 features with a Mahalanobis distance or a principal subspace lose 8 to 16 points of AUROC, and the other scores lose at most 5.5.
\section{Derivations}
\label{sec:s_derivations}

Throughout, we fix a stratum $g$ and write $r_1, \dots, r_n$ for the scores of its reference boxes and $S$ for the score of a test box of the same stratum.
The class statistics and the lines of PLUM come from other images, so they fix the score before the references enter.
As in Sec.~\ref{sec:s_details}, we treat the reference boxes as independent and leave the dependence within an image aside.

\paragraph{Proposition~\ref{prop:validity}.}
Let $l = \lfloor (n + 1)\alpha \rfloor$ and let $r_{(1)} \le \dots \le r_{(n)}$ be the sorted reference scores.
Because the count of reference scores is an integer, the rank of Eq.~\eqref{eq:rank} satisfies
\begin{equation}
  \begin{aligned}
    u(b) \le \alpha
    &\iff 1 + \big|\{ i : r_i \le S \}\big| \le (n + 1)\alpha \\
    &\iff \big|\{ i : r_i \le S \}\big| \le l - 1 \\
    &\iff S < r_{(l)},
  \end{aligned}
  \label{eq:s_event}
\end{equation}
and no box is flagged when $l = 0$.
Write $z_1, \dots, z_{n+1}$ for the reference scores and $S$ together, and $m_j$ for the $l$-th smallest of the $n$ values other than $z_j$.
Only the $l$ smallest values can lie below $m_j$, and exchangeability gives every $j$ the same probability, so
\begin{equation}
  \begin{aligned}
    P\big(u(b) \le \alpha\big)
    &= \frac{1}{n + 1} \sum_{j=1}^{n+1} P(z_j < m_j) \\
    &= \frac{1}{n + 1}\, \mathbb{E}\Big[ \sum_{j=1}^{n+1} \mathds{1}[z_j < m_j] \Big] \\
    &\le \frac{l}{n + 1} \le \alpha.
  \end{aligned}
  \label{eq:s_size}
\end{equation}
Without ties the sum equals $l$, and the first inequality becomes an equality.
Let the scores be independent draws from a continuous distribution function $F_g$, so $U_i = F_g(r_i)$ are independent and uniform on $(0, 1)$.
Let $K$ count the flagged boxes among $n_{\mathrm{ID}}$ independent ID boxes of the stratum, and write $r_{1:n}$ for the references.
Then
\begin{equation}
  \begin{aligned}
    P\big(S < r_{(l)} \mid r_{1:n}\big) &= F_g(r_{(l)}) = U_{(l)}, \\
    U_{(l)} &\sim \mathrm{Beta}(l, n + 1 - l), \\
    \mathbb{E}\big[U_{(l)}\big] &= \tfrac{l}{n + 1} \in \big[\alpha - \tfrac{1}{n + 1}, \alpha\big], \\
    K \mid r_{1:n} &\sim \mathrm{Bin}\big(n_{\mathrm{ID}}, U_{(l)}\big), \\
    K &\sim \mathrm{BetaBin}\big(n_{\mathrm{ID}}, l, n + 1 - l\big).
  \end{aligned}
  \label{eq:s_beta}
\end{equation}
One minus $U_{(l)}$ is the coverage of split conformal prediction given its calibration data~\cite{angelopoulos2024theoretical}.
Table~\ref{tab:s_exch} uses the central 95\% range of the beta-binomial distribution of $K / n_{\mathrm{ID}}$.

\paragraph{Proposition~\ref{prop:shift}.}
PLUM draws its references from the images on which the detector was trained, so we bound the false-alarm rate given the references in use.
For the references of stratum $g$, let $\hat F^{-}(t) = |\{ i : r_i < t \}| / n$, let the score of an ID test box of the stratum follow $G_g^{-}(t) = P(S < t)$, and let
\begin{equation}
  \delta_g(t_0) = \sup_{t \le t_0} \big( G_g^{-}(t) - \hat F^{-}(t) \big).
  \label{eq:s_delta}
\end{equation}
Let $l_{\max} = \lfloor (n + 1)\alpha_{\max} \rfloor$.
For $1 \le l \le l_{\max}$, Eq.~\eqref{eq:s_event} and $|\{ i : r_i < r_{(l)} \}| \le l - 1$ give
\begin{equation}
  \begin{aligned}
    P\big(u(b) \le \alpha \mid \mathrm{ID}, g, r_{1:n}\big)
    &= G_g^{-}(r_{(l)}) \\
    &\le \hat F^{-}(r_{(l)}) + \delta_g(r_{(l)}) \\
    &\le \frac{l - 1}{n} + \delta_g(r_{(l_{\max})}) \\
    &< \alpha + \delta_g(r_{(l_{\max})}),
  \end{aligned}
  \label{eq:s_shift}
\end{equation}
where the last step uses $l \le (n + 1)\alpha$, and no box is flagged when $l = 0$.
The bound holds for every $\alpha \le \alpha_{\max}$ at once.
The chain uses no property of the references, so it holds whether or not they are exchangeable with test boxes or depend on each other.
Only a shift of the ID test scores towards low values raises the false-alarm rate.
With $\alpha_{\max} = 10\%$ and the ID validation boxes in place of $G_g^{-}$, Eq.~\eqref{eq:s_delta} gives the $\delta$ of Table~\ref{tab:exch} of the main paper, and the same chain bounds the height of each stratum curve of Fig.~\ref{fig:calib} of the main paper above the diagonal.
Even exchangeable references leave $\delta_g$ above zero, because $\hat F^{-}$ deviates from the distribution of the reference scores by about $1/\sqrt{n}$.
We therefore draw $n$ reference scores and the ID scores of the stratum independently from one continuous distribution 1000 times and call $\delta$ significant when its p-value over these draws is below 0.05.

\paragraph{Eq.~\eqref{eq:power}.}
Let $F$ be the continuous distribution function of the ID scores and let $u(b) = F(S(b))$, so ID boxes have uniform ranks.
An ID box scores higher than an OOD box with score $s$ with probability $1 - F(s)$, and the threshold that keeps 95\% of the ID boxes is $F^{-1}(0.05)$.
Hence
\begin{equation}
  \begin{aligned}
    \mathrm{AUROC} &= \mathbb{E}_{\mathrm{OOD}}\big[1 - F(S(b))\big] \\
    &= 1 - \mathbb{E}_{\mathrm{OOD}}\big[u(b)\big], \\
    \mathrm{FPR95} &= P_{\mathrm{OOD}}\big(S(b) \ge F^{-1}(0.05)\big) \\
    &= P_{\mathrm{OOD}}\big(u(b) > 0.05\big) \\
    &= 1 - P_{\mathrm{OOD}}\big(u(b) \le 0.05\big).
  \end{aligned}
  \label{eq:s_auroc}
\end{equation}
With ranks within strata, $u(b) = F_{g(b)}(S(b))$ is uniform within every stratum.
For shares $\pi^{\mathrm{ID}}_g$ of ID boxes in the strata,
\begin{equation}
  \begin{aligned}
    P_{\mathrm{ID}}\big(u(b) \le a\big)
    &= \sum_g \pi^{\mathrm{ID}}_g\, P_{\mathrm{ID}}\big(u(b) \le a \mid g\big) \\
    &= \sum_g \pi^{\mathrm{ID}}_g\, a = a,
  \end{aligned}
  \label{eq:s_mixture}
\end{equation}
so ID boxes have uniform ranks over all strata too, and Eq.~\eqref{eq:s_auroc} holds for these ranks.

\paragraph{AUROC with and without strata.}
Let $\pi^{\mathrm{OOD}}_h$ be the share of OOD boxes in stratum $h$, and let $\mathcal{A}_{gh}$ be the probability that an ID box of stratum $g$ scores higher than an OOD box of stratum $h$.
One null distribution for all boxes compares every pair of an ID box and an OOD box.
Ranks within strata compare each OOD box only with ID boxes of its own stratum through Eqs.~\eqref{eq:s_auroc} and~\eqref{eq:s_mixture}.
Hence
\begin{equation}
  \begin{aligned}
    \mathrm{AUROC}_{\mathrm{one}} &= \sum_{g, h} \pi^{\mathrm{ID}}_g\, \pi^{\mathrm{OOD}}_h\, \mathcal{A}_{gh}, \\
    \mathrm{AUROC}_{\mathrm{strata}} &= 1 - \mathbb{E}_{\mathrm{OOD}}\big[u(b)\big] \\
    &= \sum_h \pi^{\mathrm{OOD}}_h \Big( 1 - \mathbb{E}\big[F_h(S(b)) \mid h\big] \Big) \\
    &= \sum_h \pi^{\mathrm{OOD}}_h\, \mathcal{A}_{hh}.
  \end{aligned}
  \label{eq:s_auroc_one}
\end{equation}
Since $\sum_g \pi^{\mathrm{ID}}_g = 1$, the two values differ only through pairs across strata,
\begin{equation}
  \begin{aligned}
    &\mathrm{AUROC}_{\mathrm{one}} - \mathrm{AUROC}_{\mathrm{strata}} \\
    &\quad = \sum_h \pi^{\mathrm{OOD}}_h \sum_{g \ne h} \pi^{\mathrm{ID}}_g \big( \mathcal{A}_{gh} - \mathcal{A}_{hh} \big).
  \end{aligned}
  \label{eq:s_auroc_diff}
\end{equation}
If all strata share one distribution of ID scores, then $\mathcal{A}_{gh} = \mathcal{A}_{hh}$ for every $g$, and one threshold rejects the same share of ID boxes in every stratum.
Both the difference and the uneven false alarms therefore arise only when the ID scores differ across strata.

\paragraph{The example of Fig.~\ref{fig:toy}.}
Small and large boxes each hold half of the ID boxes and half of the OOD boxes.
The distance of ID boxes follows $\mathcal{N}(10, 1)$ for small boxes and $\mathcal{N}(0, 1)$ for large boxes, OOD boxes follow $\mathcal{N}(12, 1)$ and $\mathcal{N}(2, 1)$, and the score is the negative distance.
With $\mathcal{A}_{\mathrm{ss}} = \mathcal{A}_{\mathrm{ll}} = \Phi(\sqrt{2})$, $\mathcal{A}_{\mathrm{sl}} = \Phi(-8/\sqrt{2})$ and $\mathcal{A}_{\mathrm{ls}} = \Phi(12/\sqrt{2})$, where the first index gives the ID kind, Eq.~\eqref{eq:s_auroc_one} gives
\begin{equation}
  \begin{aligned}
    \mathrm{AUROC}_{\mathrm{strata}} &= \Phi(\sqrt{2}) = 0.921, \\
    \mathrm{AUROC}_{\mathrm{one}} &= \tfrac{1}{2}\,\Phi(\sqrt{2}) \\
    &\quad + \tfrac{1}{4}\big[ \Phi(-8/\sqrt{2}) + \Phi(12/\sqrt{2}) \big] \\
    &= 0.711.
  \end{aligned}
  \label{eq:s_toy_auroc}
\end{equation}
One threshold $t$ on the distance keeps 95\% of all ID boxes, so $\tfrac{1}{2}\,\Phi(t - 10) + \tfrac{1}{2}\,\Phi(t) = 0.95$ and $t = 11.28$.
It rejects $1 - \Phi(t - 10) = 10\%$ of the small ID boxes and $1 - \Phi(t) \approx 0$ of the large ones.
A threshold for each kind lies $1.645$ above its ID mean.
The FPR95 values follow as
\begin{equation}
  \begin{aligned}
    \mathrm{FPR95}_{\mathrm{one}}
    &= \tfrac{1}{2}\,\Phi(t - 12) + \tfrac{1}{2}\,\Phi(t - 2) \\
    &= 0.618, \\
    \mathrm{FPR95}_{\mathrm{strata}} &= \Phi(1.645 - 2) = 0.361.
  \end{aligned}
  \label{eq:s_toy_fpr}
\end{equation}

\section{Qualitative examples}
\label{sec:s_examples}

\begin{figure*}[t]
  \centering
  \includegraphics[width=\textwidth]{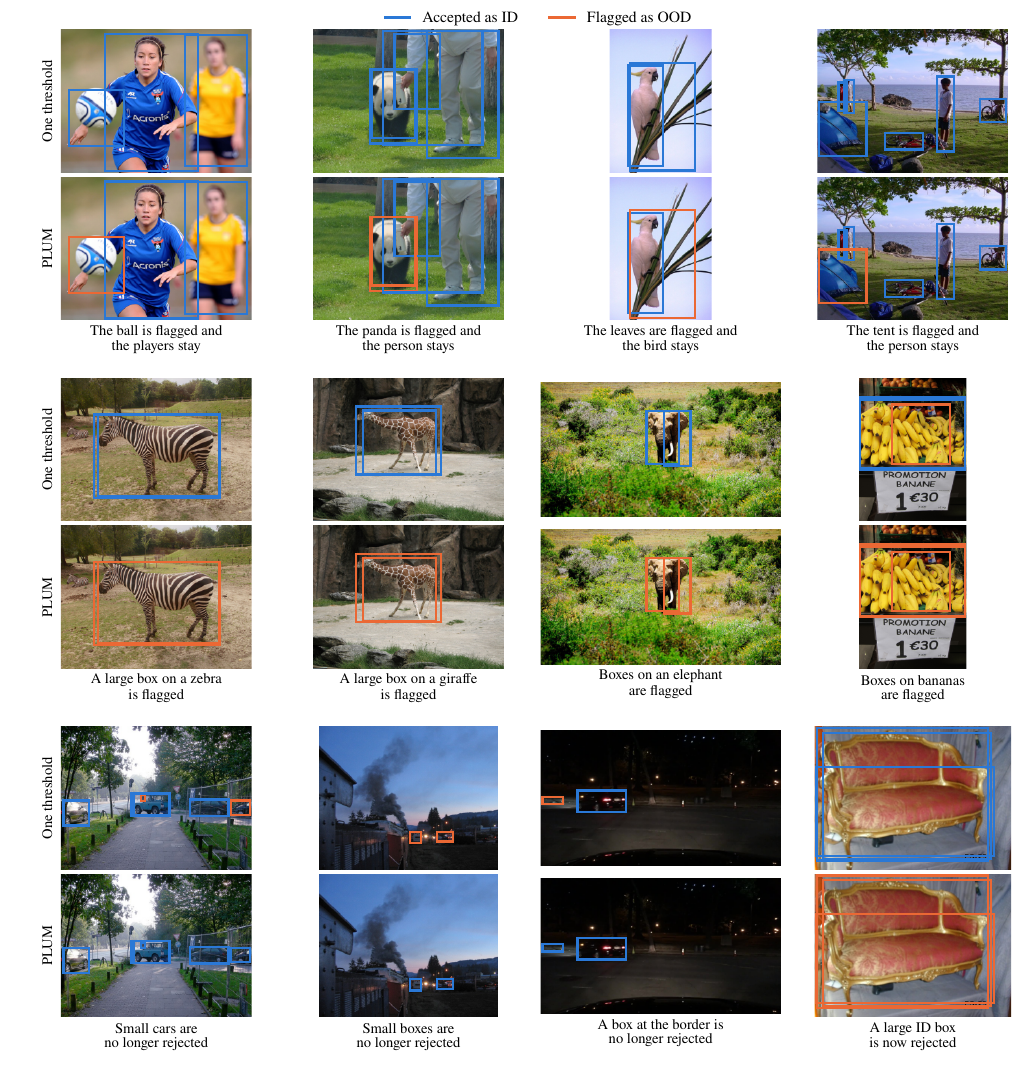}
  \caption{Decisions of one threshold for all boxes and of PLUM on single images at $\alpha = 5\%$. The examples come from the ID and OOD images of both benchmarks and from MS-COCO images that contain ID and OOD objects together, each scored by the detector of its ID dataset. The benchmark counts every detection on an ID image as ID. These images illustrate each case and are not a sample.}
  \label{fig:s_examples}
\end{figure*}

Figure~\ref{fig:s_examples} extends Fig.~\ref{fig:examples} of the main paper to whole images.
Each example shows the decisions of one threshold for all boxes above those of PLUM on the same image at $\alpha = 5\%$, and the color of a box gives its decision.

In the first row, PLUM flags an unknown object and keeps the known objects of the same image.
Some of these images come from MS-COCO and contain objects of both kinds.
The others are ID images of the benchmark, which counts the detection on the unknown object as ID.
In the second row, PLUM flags large boxes on unknown objects that one threshold accepts, as in the example of Fig.~\ref{fig:toy} of the main paper.
In the third row, small boxes and boxes at the image border are no longer rejected, while a large ID box now is.
PLUM thus does not remove false alarms but spreads them evenly over the kinds of box, as Sec.~\ref{sec:exp_size} of the main paper measures.

\end{document}